\documentclass[11pt, letterpaper]{article}
\usepackage[table]{xcolor}
\usepackage{amsmath}
\usepackage{amssymb}
\usepackage{amsthm}
\usepackage{graphicx}
\usepackage{hyperref}
\usepackage{url}
\usepackage{algorithm}
\usepackage{algpseudocode}

\usepackage{amsmath}
\usepackage{caption}
\usepackage{subcaption}
\usepackage{booktabs}
\usepackage{authblk}
\usepackage{appendix}
\usepackage[utf8]{inputenc}
\usepackage[T1]{fontenc}
\usepackage[english]{babel}
\usepackage{geometry}
\usepackage{xcolor}
\usepackage{fancyhdr}
\usepackage{microtype}
\usepackage{parskip}
\usepackage{tabularx}
\usepackage{enumitem}
\usepackage{multirow}
\usepackage{array}
\usepackage{booktabs}
\usepackage{ragged2e}
\usepackage{colortbl}
\usepackage{longtable}
\usepackage{adjustbox}
\usepackage{cite}

\usepackage{makecell}

\hypersetup{
    colorlinks=true,
    linkcolor=blue,
    filecolor=magenta,      
    urlcolor=cyan,
    citecolor=red,
}

\definecolor{headercolor}{RGB}{0, 50, 100}
\title{\Large ChatRadio-Valuer: A Chat Large Language Model for Generalizable Radiology Report Generation Based on Multi-institution and Multi-system Data}

\date{}

\newcommand*\samethanks[1][\value{footnote}]{\footnotemark[#1]}
\author[1]{Tianyang Zhong \thanks{Co-first authors.}}
\author[2,3,4]{Wei Zhao \samethanks}

\author[5]{Yutong Zhang \thanks{Co-second authors.}}
\author[6]{Yi Pan \samethanks}
\author[1]{Peixin Dong \samethanks}
\author[1]{Zuowei Jiang \samethanks}
\author[8]{Xiaoyan Kui \samethanks}
\author[2]{Youlan Shang \samethanks}

\author[1]{Li Yang}
\author[1]{Yaonai Wei}
\author[2]{Longtao Yang}
\author[1]{Hao Chen}
\author[7]{Huan Zhao}
\author[9,10]{Yuxiao Liu}
\author[1]{Ning Zhu}
\author[18]{Yiwei Li}
\author[2]{Yisong Wang}
\author[11]{Jiaqi Yao}
\author[12]{Jiaqi Wang}
\author[2,13]{Ying Zeng}
\author[14]{Lei He}
\author[15]{Chao Zheng}
\author[16]{Zhixue Zhang}
\author[17]{Ming Li}
\author[18]{Zhengliang Liu}
\author[18]{Haixing Dai}
\author[18]{Zihao Wu}
\author[19]{Lu Zhang}

\author[12]{Shu Zhang}
\author[1]{Xiaoyan Cai}
\author[1]{Xintao Hu}
\author[1]{Shijie Zhao}
\author[7]{Xi Jiang}
\author[5]{Xin Zhang}
\author[20]{Xiang Li}
\author[19]{Dajiang Zhu}
\author[1]{Lei Guo}
\author[9,21,22]{Dinggang Shen}
\author[1]{Junwei Han}
\author[18]{Tianming Liu}
\author[2,3]{Jun Liu \thanks{Co-corresponding authors. Second corresponding author. Email: junliu123@csu.edu.cn}}
\author[1]{Tuo Zhang \thanks{Co-corresponding authors. First corresponding author. Email: zhangtuo.npu@gmail.com}}

\affil[1]{School of Automation, Northwestern Polytechnical University, Xi'an 710072, China}
\affil[2]{The Second Xiangya Hospital, Central South University, Changsha 410011, China}
\affil[3]{Clinical Research Center for Medical Imaging in Hunan Province, Changsha, China}
\affil[4]{Institute of Biomedical and Health Engineering, Shenzhen Institutes of Advanced Technology, Chinese Academy of Sciences, Shenzhen 518055, China}
\affil[5]{Institute of Medical Research, Northwestern Polytechnical University, Xi'an 710072, China}
\affil[6]{Glasgow College, University of Electronic Science and Technology of China, Chengdu 611731, China }
\affil[7]{School of Life Science and Technology, University of Electronic Science and Technology of China, Chengdu 611731, China}
\affil[8]{School of Computer Science and Engineering, Central South University, Changsha 410083, Hunan Province, People's Republic of China}
\affil[9]{School of Biomedical Engineering, ShanghaiTech University, Shanghai 201210, China}
\affil[10]{Lingang Laboratory, Shanghai, 200031, China}	
\affil[11]{Imaging Center, the Second Affiliated Hospital of Xinjiang Medical University, Urumuqi 830000, China}
\affil[12]{School of Computer Science, Northwestern Polytechnical University, Xi'an 710072, China}
\affil[13]{Department of Radiology, Xiangtan Central Hospital, Xiangtan 411199, China}
\affil[14]{Department of Radiology, Yueyang Central Hospital City, Yueyang 414000, China}
\affil[15]{Department of Radiology, The First People's Hospital of Changde City, Changde 415003, China}
\affil[16]{Department of Radiology, The First Hospital of Hunan University of Chinese Medicine, Changsha 410021, China}
\affil[17]{Department of Radiology, Huadong Hospital Affiliated to Fudan University, Shanghai 200040, People's Republic of China}
\affil[18]{School of Computing, University of Georgia, GA, USA}
\affil[19]{Department of Computer Science and Engineering, University of Texas at Arlington, TX, USA}
\affil[20]{Department of Radiology, Massachusetts General Hospital and Harvard Medical School, MA, USA}
\affil[21]{Shanghai United Imaging Intelligence Co., Ltd.}
\affil[22]{Shanghai Clinical Research and Trial Center}

\begin{document}

\maketitle

\begin{abstract}
Radiology report generation, as a key step in medical image analysis, is critical to the quantitative analysis of clinically informed decision-making levels. However, complex and diverse radiology reports with cross-source heterogeneity pose a huge generalizability challenge to the current methods under massive data volume, mainly because the style and normativity of radiology reports are obviously distinctive among institutions, body regions inspected and radiologists. Recently, the advent of large language models (LLM) offers great potential for recognizing signs of health conditions. To resolve the above problem, we collaborate with the Second Xiangya Hospital in China and propose ChatRadio-Valuer based on the LLM, a tailored model for automatic radiology report generation that learns generalizable representations and provides a basis pattern for model adaptation in sophisticated analysts' cases. Specifically, ChatRadio-Valuer is trained based on the radiology reports from a single institution by means of supervised fine-tuning, and then adapted to disease diagnosis tasks for human multi-system evaluation (i.e., chest, abdomen, muscle-skeleton, head, and maxillofacial $\&$ neck) from six different institutions in clinical-level events. The clinical dataset utilized in this study encompasses a remarkable total of \textbf{332,673} observations. From the comprehensive results on engineering indicators, clinical efficacy and deployment cost metrics, it can be shown that ChatRadio-Valuer consistently outperforms state-of-the-art models, especially ChatGPT (GPT-3.5-Turbo) and GPT-4 et al., in terms of the diseases diagnosis from radiology reports. ChatRadio-Valuer provides an effective avenue to boost model generalization performance and alleviate the annotation workload of experts to enable the promotion of clinical AI applications in radiology reports.

\end{abstract}

\section{Introduction}
Radiology exists as an essential department in hospitals or medical institutions, which substantially facilitates the progress of the detection, diagnosis, prediction, evaluation and follow-up for most of diseases. Radiology reports provide a comprehensive evaluation for different diseases, assisting clinicians in making decisions \cite{doi:10.1148/radiol.231259,choy2018current,zhang2020radiology,prokop2020co,xue2018multimodal}. The natural process of generating a radiology report is that the radiologists conclude an impression after comprehensively describing the positive and negative image findings of a specific examination \cite{yuan2019automatic,J2,liu2019clinically,nie2018medical} (i.e., chest CT and abdominal CT). Considering the tremendous amounts of radiology examinations, especially in China, it places a heavy burden on radiologists to write radiology reports in daily work \cite{lecler2023revolutionizing,doi:10.1148/rg.2015150023,monshi2020deep}. Besides, the style and normativity of radiology reports are variable among individuals and institutions, due to the different writing habits and education. This may confuse the patients if they receive different descriptions from many institutions and also limit the communication among different radiologists. Therefore, an efficient and standard way to generate the impression of radiology report is pursued by radiologists.

Driven by the extensive clinical demands, automatic generating radiology reports becomes a new hotspot and promising direction \cite{pons2016natural,mozayan2021practical,J1,J3,J4,J5,J6,J7,J8,J9,J10}. In this context, natural language processing (NLP) strategies, which are widely used in non-medical areas, are recently adopted to tackle this issue \cite{luo2020review}. With the aid of NLP techniques, the reporting time can be obviously reduced, resulting in the improvement of work efficiency. Another potential advantage of NLP in generating radiology reports is that the impression of radiology reports could be more structured and complete, further reducing error rates and facilitating the mutual communication from different institutions. In general, there are two main research directions: model-driven methods and data-driven methods. 

Many previous studies on radiology report generation have followed model-driven methods \cite{J1,J2,J3,J4,J5}. Jing et al. \cite{J1} proposed a co-attention network to produce full paragraphs for automatically generating reports and demonstrated the effectiveness of their proposed methods on publicly available datasets. Chen et al. \cite{J2} proposed to generate radiology reports via a memory-driven transformer, which addressed what conventional image captioning methods are inefficient (and inaccurate) for generating long and detailed radiology reports. Chen et al. \cite{J3} proposed that incorporating memory into both the encoding process and decoding process can further enhance the generation ability of the transformer. Wang et al., \cite{J4} proposed a cross-modal network to facilitate cross-modal pattern learning, where the cross-modal prototype matrix is initialized with prior information and an improved multi-label contrastive loss is proposed to promote the cross-modal prototype learning. Wang et al., \cite{J5} proposed a self-boosting framework, where the main task of report generation helps learn highly text-correlated visual features, and an auxiliary task of image-text matching facilitates the alignment of visual features and textual features.

Other researchers deem that data is significant for deep learning based automatic report generation, and their research interests focus on data-driven methods \cite{J6,J7,J8,J9,J10}. Liu et al. \cite{J6} introduced three modules to utilize prior knowledge and posterior knowledge from radiology data for alleviating visual and textual data bias, and showed that their model outperforms previous methods on both language generation metrics and clinical evaluation. Inspired by curriculum learning, Nooralahzadeh et al. \cite{J7} extracted global concepts from the radiology data and utilized them as transition text from chest radiographs to radiology reports (i.e., image-to-text-to-text). To alleviate the data bias and make the best use of available data, Liu et al. \cite{J8} proposed a competence-based multimodal curriculum learning framework, where each training instance was estimated and current models were evaluated and then the most suitable batch of training instances were selected considering current model competence. To make full use of limited data, Yan et al. \cite{J9} developed a weakly supervised approach to identify “hard” negative samples from radiology data and assign them with higher weights in training loss to enhance sample differences. Yuan et al. \cite{J10} introduced a sentence-level attention mechanism to fuse multi-view visual features and extracted medical concepts from radiology reports to fine-tune the encoder for extracting the most frequent medical concepts from the x-ray images.

The above methods have made great progress in radiology report generation. However, there are still some limitations. Firstly, the current expansion of radiology data is largely contributed to the practice of sharing across multiple medical institutions, which leads to a complex data interaction, thus requiring insightful and controlled analysis. Secondly, the development of modern radiology imaging leads to data complexity increasing.  Thirdly, in radiology report generation, NLP researchers have paid attention to image caption \cite{vinyals2016show}, which has demonstrated its effectiveness. Actually, the transfer properties of these models to radiology report generation encounter the model generalization problem. Model generalization requires a trained model to transfer its generation capacity to the target domain and produce accurate output for previously unseen instances. In radiology report generation, this requires generating clinically accurate reports for various medical subjects. While image caption has exhibited remarkable effectiveness in general text generation tasks, the complexities of medical terms and language introduce hindrances to model generalization.

As one of the most influential artificial intelligence (AI) products today \cite{dai2023chataug}, LLMs, known as generative models, provide a user-friendly human-machine interaction platform that brings the powerful capabilities of large language models to the public and has been rapidly integrated into various fields of application \cite{dai2023chataug,liu2023summary,wang2023chatcad,ma2023impressiongpt,chowdhery2022palm,longpre2023flan,scao2022bloom,zhang2022opt,JJ1,JJ2,JJ3,JJ4,JJ5,JJ6,JJ7,JJ8,JJ9}. Liu et al. \cite{JJ8} proposed Radiology-GPT, which uses instruction tuning approach to fine-tune Alpaca \cite{alpaca} for mining radiology domain knowledge, and prove its superior performance on reports summarization task. Based on Llama2, Liu et al. \cite{JJ9} further presented Radiology-Llama2 using instruction tuning to fine-tune Llama2 on radiology reports. However, they both ignore the fact that the style and normativity of radiology reports vary among individuals and institutions with writing habits and education. Therefore, they are difficult to achieve satisfactory results when training medical LLMs.


Inspired by large language models, we propose ChatRadio-Valuer by fine-tuning Llama2 on large-scale, high-quality instruction-following data in the radiology domain. ChatRadio-Valuer inherits robust language understanding ability across various domains and genres, coupled with complex reasoning and diverse generation abilities. Additionally, it learns high-level domain-specific knowledge during training in the radiology domain. This addresses the heterogeneity gaps among institutions and enables accurate radiology report generation. Extensive experiments are conducted with real radiology reports from a clinical pipeline in the Second Xiangya Hospital of China and the experimental results highlight the superiority of our method. The main contributions are summarized as follows:

\begin{itemize}

\item To the best of our knowledge, a complete full-stack solution for clinical-level radiology report generation based on multi-institution and multi-system data is developed to obtain desirable performance results for the first time. The solution significantly outperforms the state-of-the-art counterparts, benefiting performance analysts of multiple data sources in radiology.
\item An effective ChatRadio-Valuer framework is proposed that can automatically utilize radiology domain knowledge for cross-institution adaptive radiology report generation based on single-institution samples, which can provide a fundamental scheme to boost model generalization performance from radiology reports.
\item We implement our framework and conduct substantial use cases on clinical utilities among six different institutions. Through the cases, we obtain valuable insights into radiology experts, which are beneficial to alleviate the annotation workload of experts. This opens the door to bridge LLMs' domain adaptation application and radiology performance evaluation. 
\end{itemize}


\begin{figure*}[!htbp]
\centering
\includegraphics[width=6.5in,height=!,keepaspectratio]{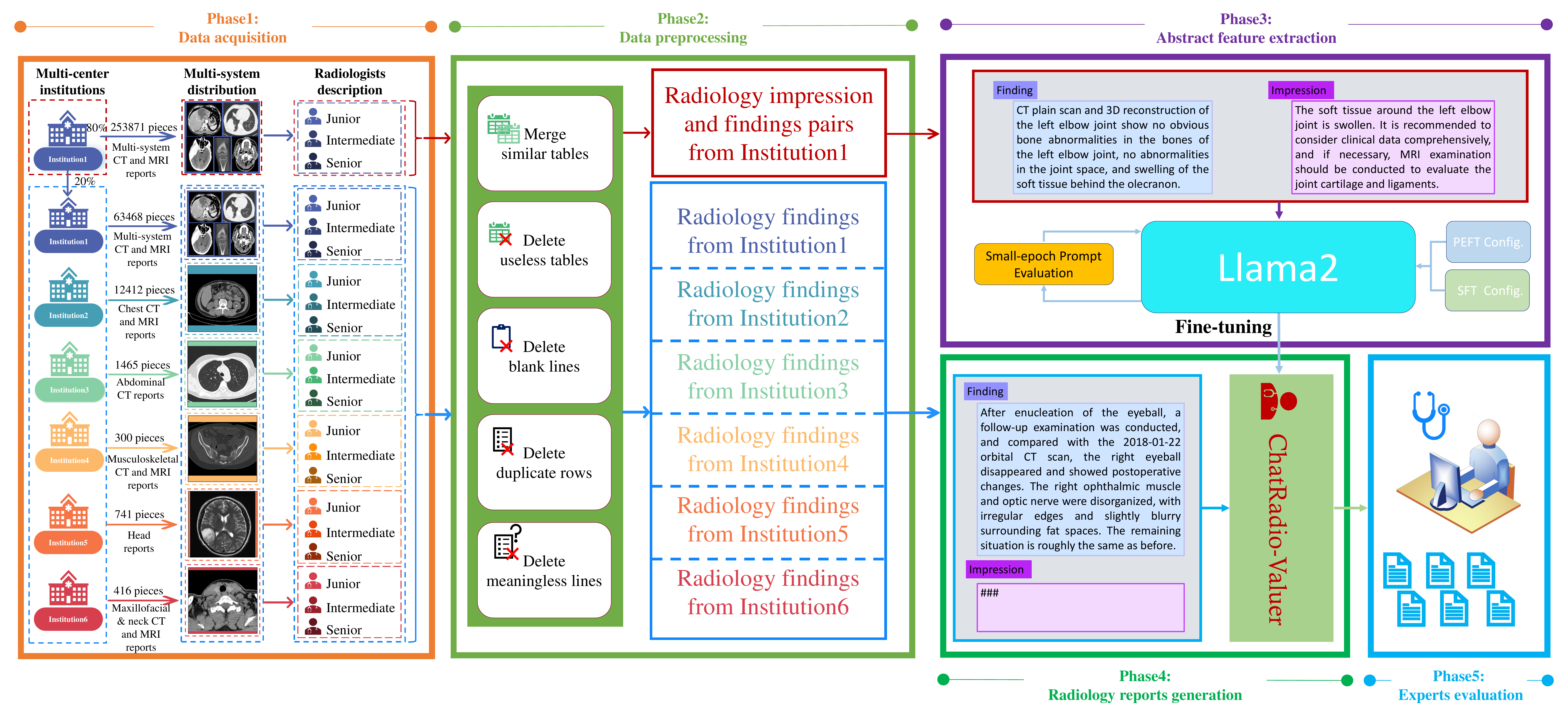}%
\caption{Overall framework of the proposed method for radiology report generation. Multi-institution and multi-system clinical radiology reports are acquired in phase 1. Systematic data preprocessing is implemented and then synthesizes the samples into high-quality prompts in phase 2. The generalizable advanced features are extracted and applied for clinical utilities in phase 3 \& phase 4. The comprehensive evaluations on ChatRadio-Valuer's efficacy are executed in phase 5.}\label{fig:overflow}
\end{figure*}

\section{Clinical Background and Data}
\subsection{Collaboration with the Second Xiangya Hospital of China}\label{sec:Collaboration with the Second Xiangya Hospital of China}

We have collaborated with the Second Xiangya Hospital of China. During the collaboration, we utilize the platform (the phase 1 in Figure \ref{fig:overflow}) for data acquisition that enables physicians to gather and analyze medical data. The platform consists of three components: multi-center institutions inspection, multi-system analysis, and radiologists' description. The ultimate goal is to propose a unified and generalizable framework, which can simulate radiology experts to execute disease diagnoses across multiple institutions and multiple systems, in order to improve clinical efficiency and reduce the workload of relevant staff in processing massive data.

Many factors have brought great challenges to this task. Specifically, in the first component of the platform, it describes that data acquisition comes from six different institutions, and the quality and quantity are different to varying degrees. The main reasons are the constraints of types of equipment and observable conditions among them. The specific materials will be introduced in Section \ref{sec:Data Description}. Next, the second component contains two message formats, such as CT and MRI modalities on the chest, abdomen, muscle-skeleton, head, and maxillofacial $\&$ neck systems. The prominent dissimilarity within this dataset arises primarily from significant variations in data source distributions between the two modalities, contingent upon the specific human body locations inspected. Specifically, Institution 1 contributes data for five complete systems, while institutions 2 through 6 provide data for individual human body systems, namely the chest, abdomen, musculoskeletal, head, and maxillofacial $\&$ neck systems. Then, during the third process, the information carrier from image to text conversion is performed by different levels of clinicians including junior, intermediate, and senior radiologists. Due to the different writing habits and education, the style and normativity of radiology reports are obviously distinctive among radiologists. 

In brief, multiple aspects, from institutions, multiple systems of human body, and radiologists, bring great dilemmas to the assignment, which may confuse the patients if they receive different descriptions from many institutions and also limit the free communication among different radiologists. The proposed framework named ChatRadio-Valuer is described in Section \textbf{ \ref{Framework}}.

\subsection{Data Description} \label{sec:Data Description}

\begin{figure}[!t]
    \centering    \includegraphics[width=1\textwidth,height=!,keepaspectratio]{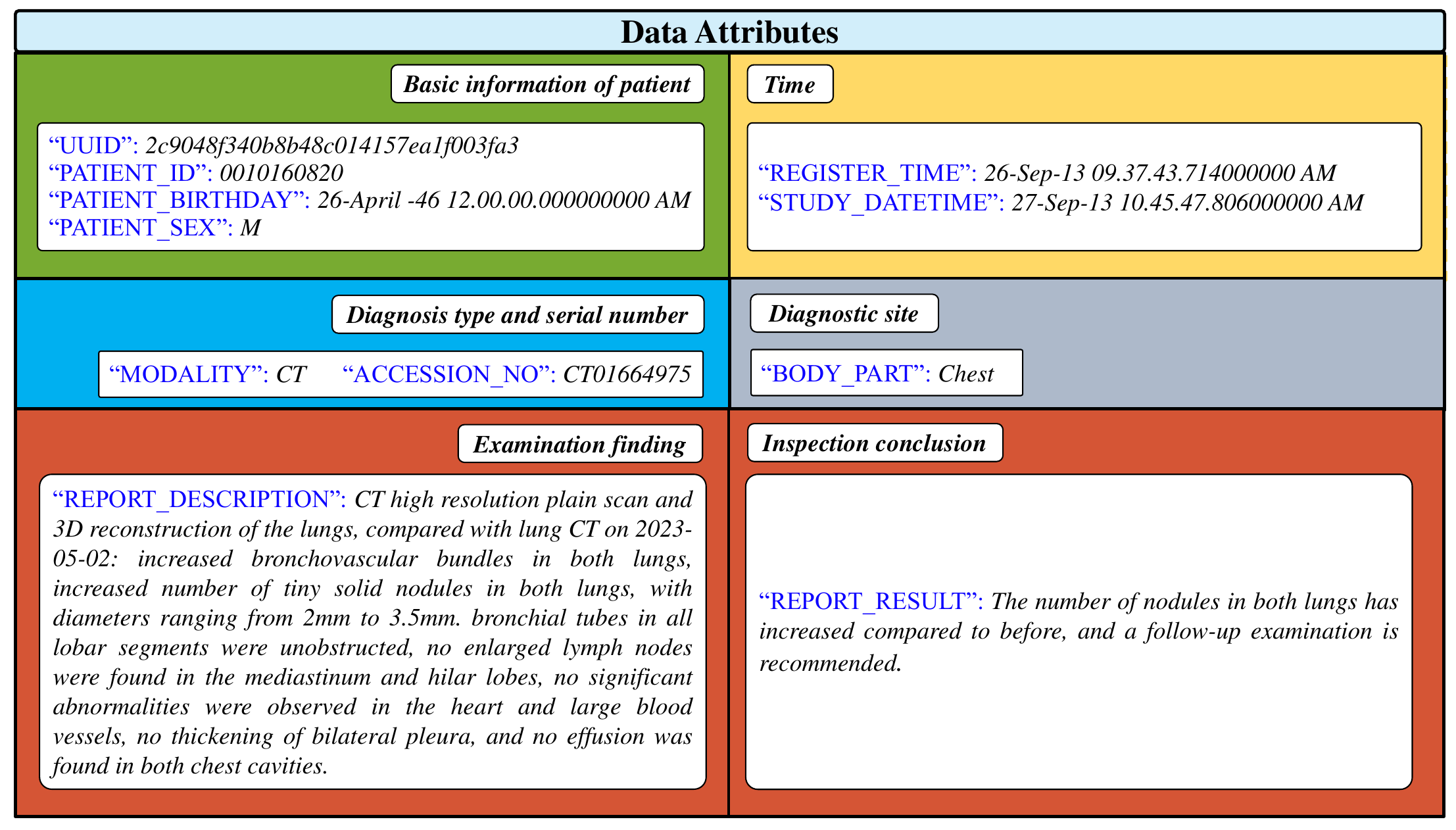}
    \caption{A radiology report example on its attributes. These attributes are manually diagnosed and described by radiologists at different levels, among which there are obvious variations in styles.}
    \label{fig:main1}
\end{figure}

\begin{table}[!t]
\centering
\caption{The statistical analysis encompasses comprehensive information regarding the experimental dataset, including both the training and test sets, along with detailed statistical insights derived from data obtained from six distinct institutions.}
\label{fig:statistical information}
\resizebox{1.0\textwidth}{!}
{
\normalsize
\begin{tabular}{l c c c c c c c c c c}
\noalign{\smallskip}
\toprule[2pt]
\noalign{\smallskip}
\multirow{2}{*}{\makebox[0.2\textwidth][l]{Variable}} &
\multicolumn{3}{c}{\makebox[0.1\textwidth][c]{Training Set}} &
\multicolumn{6}{c}{\makebox[0.1\textwidth][c]{Test Set}} &
\multicolumn{1}{c}{\makebox[0.1\textwidth][c]{Total}}\\
\cmidrule(lr){2-4} \cmidrule(lr){5-10} \cmidrule(lr){11-11}
 & 
 \multicolumn{3}{c}{\makebox[0.1\textwidth][c]{\scriptsize \makecell{Institution 1 \\ (80$\%$)}}} &
 \makebox[0.1\textwidth][c]{\scriptsize \makecell{Institution 1 \\ (20$\%$)}} &
 \makebox[0.1\textwidth][c]{\scriptsize Institution 2} &
 \makebox[0.1\textwidth][c]{\scriptsize Institution 3} &
 \makebox[0.1\textwidth][c]{\scriptsize Institution 4} &
 \makebox[0.1\textwidth][c]{\scriptsize Institution 5} &
 \makebox[0.1\textwidth][c]{\scriptsize Institution 6} &
 \makebox[0.1\textwidth][c]{\scriptsize All Set} \\
\noalign{\smallskip}
\hline
\noalign{\smallskip}
\rowcolor{cyan!10}
No.of pieces &\qquad253871 & & &63468 &12412 &1465 &300 &741 & 416 &332673\\
\noalign{\smallskip}
Age(y) &\qquad0-103 & & &0-101 &0-96 &9-95 &1-92 &0-100 & 0-91 &0-103\\
\noalign{\smallskip}
\rowcolor{cyan!10}
Sex & & & & & & & & &  &\\
\noalign{\smallskip}
\quad Female &\qquad 119185 & & &28993 &7320 &620 &133 &382 & 225 &156858\\
\noalign{\smallskip}
\rowcolor{cyan!10}
\quad Male &\qquad 134686 & & &34475 &5092 &845 &167 &359 & 191 &175815\\
\noalign{\smallskip}
Modality & & & & & & & & &  &\\
\noalign{\smallskip}
\rowcolor{cyan!10}
\quad CT &\qquad 210487 & & &49488 &12397 &1465 &238 &543 & 346 &274964\\
\noalign{\smallskip}
\quad MRI &\qquad 43384 & & &13980 &15 &0 &62 &198 & 70 &57709\\
\noalign{\smallskip}
\rowcolor{cyan!10}
System & &  & & & & & & &  &\\
\noalign{\smallskip}
\quad Chest &\qquad 75277 & & &18820 &12412 &0 &0 &0 & 0 &106509\\
\noalign{\smallskip}
\rowcolor{cyan!10}
\quad Abdomen &\qquad 51640 & & &12910 &0 &1465 &0 &0 & 0 &66015\\
\noalign{\smallskip}
\quad Muscle-skeletion &\qquad 36874 & & &9218 &0 &0 &300 &0 & 0 &46392\\
\noalign{\smallskip}
\rowcolor{cyan!10}
\quad Head &\qquad 55922 & & &13980 &0 &0 &0 &741 & 0 &70643\\
\noalign{\smallskip}
\quad Maxillofacial $\&$ neck &\qquad 34158 & & &8540 &0 &0 &0 &0 & 416 &43114\\
\noalign{\smallskip}
\bottomrule[2pt]
\noalign{\smallskip}
\end{tabular}
}
\vspace{-6pt}
\end{table}


In this multi-institution modeling study, the radiology reports, including basic information, description and impression, were searched and downloaded in the report system from six institutions. In detail, the specific data attributes of a radiology report example are shown in Figure \ref{fig:main1}. And a comprehensive data distribution is illustrated in Table \ref{fig:statistical information}. In the Second Xiangya hospital dataset (Dataset SXY), 317339 radiology reports from five subgroups (94097 chest reports, 64550 abdominal reports, 46092 musculoskeletal reports, 69902 head reports, and 42698 maxillofacial $\&$ neck reports) were exported from 2012 year to 2023 year. These data were used for model development and internal testing. Other five datasets were collected for external testing, from the Huadong Hospital dataset (Dataset External 1, DE1, chest reports), Xiangtan Central Hospital (Dataset External 2, DE2, abdominal reports), The First Hospital of Hunan University of Chinese Medicine (Dataset External 3, DE3, musculoskeletal reports), The First People’s Hospital of Changde City (Dataset External 4, DE4, head reports), Yueyang Central Hospital City (Dataset External 5, DE5, maxillofacial $\&$ neck reports). 
They are denoted as from Institution 1 to Institution 6 respectively. This retrospective study was approved by The Second Xiangya Hospital, Institutional Review Board (Approval NO. LYF2023084), which waived the requirement for informed consent.

\subsection{Data Preprocessing} \label{sec:Data Preprocessing}
In the overall framework of ChatRadio-Valuer, the quality of radiology reports plays a critical role in facilitating the method's performance. Hence, the data preprocessing phase, encompassing the multi-system \& multi-institution data refinement and prompt generation, assumes a pivotal role in ensuring the model's efficacy and dependability when dealing with extensive domain-specific radiology reports originating from diverse institutions and systems. To achieve this objective, the operations proposed in this study are executed as indicated in Algorithm \ref{alg:prep}, aiming at a particular emphasis on mitigating improper interference caused by value and report title repetition, sheet inconsistencies and text irrelevance. The specific details are as follows:

\begin{algorithm}[H]
\caption{Training Preparation.}\label{alg:prep}
\begin{algorithmic}[1]
\Require
Expert-curated prompt templates (system description and instruction in Figure \ref{fig:promptgeneration}) $\boldsymbol{X}_{pt}=\left\{\boldsymbol{t}_{1},\boldsymbol{t}_{2},\ldots,\boldsymbol{t}_{5}\right\}$; 
Multi-institution reports $\boldsymbol{X}_{c}=\left\{\boldsymbol{c}_{1},\boldsymbol{c}_{2},\ldots,\boldsymbol{c}_{6}\right\}$, the $k^{\mathrm{th}}$ system data within the $j^{th}$-institution $\boldsymbol{c}^{k}_{j}=\left\{(\boldsymbol{x}^{k}_{j1},\boldsymbol{y}^{k}_{j1}),(\boldsymbol{x}^{k}_{j2},\boldsymbol{y}^{k}_{j2}),\ldots,(\boldsymbol{x}^{k}_{jl},\boldsymbol{y}^{k}_{jl})\right\}$; 
Specialized meaningless word set $\boldsymbol{M}$
\Ensure
Cleaned and synthesized test-institution data $\boldsymbol{X}_{test}$ and prompts $\boldsymbol{X}_{prompt}$ which contains training prompts $\boldsymbol{X}^{tr}_{prompt}$ and evaluation prompts $\boldsymbol{X}^{eval}_{prompt}$; Training labels $\boldsymbol{L}_{tr}$; Evaluation labels $\boldsymbol{L}_{eval}$
\State $\boldsymbol{X}^{n}_{c} = rand(\sum_{m=1}^{6} \left\{(\boldsymbol{x}^{n}_{m1},\boldsymbol{y}^{n}_{m1}),(\boldsymbol{x}^{n}_{m2},\boldsymbol{y}^{n}_{m2}),\ldots,(\boldsymbol{x}^{n}_{ml},\boldsymbol{y}^{n}_{ml}))\right\}$ \textbf{ for} $ n=1,\ldots,k$
\State $\boldsymbol{X}_{all}=\sum_{m=1}^{k}\boldsymbol{X}^{m}_{c}$
\State $\boldsymbol{X}_{train}, \boldsymbol{X}_{test}=ExtractSet(\boldsymbol{X}_{all})$
\State $\boldsymbol{X}_{tr},\boldsymbol{X}_{eval} = 0.8\boldsymbol{X}_{train},0.2\boldsymbol{X}_{train}$
\For{$\boldsymbol{n}:=$ \textbf{tr,eval}} 
\State $\boldsymbol{X}^{cln}_{n} = SynMultiSheet(RemoveRepTitle(RemoveRepVal(\boldsymbol{X}_{n})))$
\While{$m \subset \boldsymbol{M} \textbf{ exist}$} 
\State $\boldsymbol{X}^{cln}_{n}=DeleteSub(\boldsymbol{m},\left\{(\boldsymbol{x}_{n1},\boldsymbol{y}_{n1}),(\boldsymbol{x}_{n2},\boldsymbol{y}_{n2}),\ldots,(\boldsymbol{x}_{nl^{'}},\boldsymbol{y}_{nl^{'}})\right\})$
\EndWhile
\State $\boldsymbol{X}^{n}_{prompt}=SynthesizePrompt(\boldsymbol{X}_{pt}, \left\{\boldsymbol{x}_{n1},\boldsymbol{x}_{n2},\ldots,\boldsymbol{x}_{nl^{'}}\right\})=\left\{\boldsymbol{sp}^{n}_{1},\boldsymbol{sp}^{n}_{2},\ldots,\boldsymbol{sp}^{n}_{i}\right\}$
\EndFor
\State $\boldsymbol{L}_{tr},\boldsymbol{L}_{eval}=\left\{\boldsymbol{y}_{tr1},\boldsymbol{y}_{tr2},\ldots,\boldsymbol{y}_{trl^{'}}\right\},\left\{\boldsymbol{y}_{e1},\boldsymbol{y}_{e2},\ldots,\boldsymbol{y}_{el^{''}}\right\}$
\State $\boldsymbol{X}_{prompt} = \left\{\boldsymbol{X}^{tr}_{prompt}, \boldsymbol{X}^{eval}_{prompt}\right\}$
\State \Return $\boldsymbol{X}_{test}, \boldsymbol{X}_{prompt}, \boldsymbol{L}_{tr}, \boldsymbol{L}_{eval}$
\end{algorithmic}
\end{algorithm}

1) \textbf{Data Synthesis:} We aggregate data from multiple systems, group them by institution, and randomly process them according to their belonging centers, obtaining a comprehensive dataset $\boldsymbol{X}_{all}$ covering all the reports. Then we follow the sequence of these institutions, extracting the training set $\boldsymbol{X}_{train}$ and the test set $\boldsymbol{X}_{test}$ according to the index of centers. In detail, the data from Institution 1 are randomly selected to fine-tune ChatRadio-Valuer, and the other institutions are utilized to test the performance of ChatRadio-Valuer. The entire training dataset are then split into an 80\% training set and a 20\% evaluation set. Data in $\boldsymbol{X}_{test}$ are exclusively reserved for evaluation purposes for simplicity.

2) \textbf{Data Cleaning:} For each dataset, a rigorous data cleaning process is conducted from three different perspectives: addressing repetition, rectifying inconsistencies, and filtering out text insignificance. To tackle repetition issues, manual searches are conducted to identify and eliminate extraneous elements related to values and report titles. This result in a curated dataset featuring unique values in a standardized format. In addressing inconsistencies arising from multiple sheets from different institutions, meticulous human scrutiny and manual consolidation are employed. Regarding text insignificance, a collaborative effort between experts from the Second Xiangya Hospital of China and our team led to the creation of a corpus containing non-essential terms, deemed too eclectic for LLMs to comprehend but crucial for clinical applications. Leveraging this corpus, a significant portion of irrelevant data is removed, yielding a refined dataset optimized for prompt generation. We combine this dataset and expert-curated templates $\boldsymbol{X}_{pt}$ to eventually construct high-quality prompts $\boldsymbol{X}_{prompt}$ consisting of both training prompts $\boldsymbol{X}^{tr}_{prompt}$ and evaluation prompts $\boldsymbol{X}^{eval}_{prompt}$, as described in the Section \textbf{\ref{Framework}}.

\section{Framework} \label{Framework}
In this section, we first define the problem of radiology report generation in this work and present an overview of ChatRadio-Valuer. Then, we introduce the implementation details of ChatRadio-Valuer.

\subsection{Problem Formulation}
Due to factors of variations in physicians' professional skills, differences in the operation of medical equipment across different hospitals, and variances in organ structures, radiology medical data often exhibit inconsistencies in both format and quality. These factors can be collectively referred to as heterogeneity in healthcare data.

In the testing task, the input is defined as \( \text{Input} = \{f, X_{\text{test}}\} \), where $f$ represents the fine-tuned model, and $X_{\text{test}}$ belongs to the set \{$X_1, X_2, X_3, X_4, X_5, X_6$\}, which denotes the data from six different institutions. In a single sample, $x_m$ represents "REPORT\_DESCRIPTION", while $y_m$ denotes "REPORT\_RESULT".Each subset $X_i$ can be further divided based on the system:
\begin{align*}
X_{\text{test}} &= X_i^j = \{(x_1, y_1), (x_2, y_2), \ldots, (x_m, y_m)\} \\
j &=
\begin{cases}
\{ \text{chest} \} & \text{if } i = 2 \\
\{ \text{abdomen} \} & \text{if } i = 3 \\
\{ \text{muscle-skeleton} \} & \text{if } i = 4 \\
\{ \text{head} \} & \text{if } i = 5 \\
\{ \text{maxillofacial $\&$ neck} \} & \text{if } i = 6 \\
\{ \text{chest, abdomen, muscle-skeleton, head, maxillofacial $\&$ neck} \} & \text{if } i = 1
\end{cases} \\
X_i &= X_i^j \quad \text{for } i=1,2,3,4,5,6
\end{align*}
Formally, the mathematical expression for the testing process can be represented as:

\begin{equation} \label{eq1}
\max_{f} \sum_{X_i^j \in X_{\text{test}}} \text{Performance}(f, X_i^j)
\end{equation}

where $f$ represents the fine-tuned model, which needs to be evaluated in the testing task; $X_i^j$ denotes a sample set from the $X_{\text{test}}$ collection, corresponding to different medical departments or system divisions; $\text{Performance}(f, X_i^j)$ is the metrics used to measure the performance of model $f$  on the sample set $X_i^j$. In Section \textbf{\ref{sec:PerformanceEvaluation}}, the similarity metrics R-1, R-2, and R-L are used as indicators.


Building upon this, we conduct experiments to assess the model's performance from the following three aspects:

\begin{itemize}
    \item Selecting $X_{\text{test}} = \{X_1, X_2, X_3, X_4, X_5, X_6\}$, we validate the model's cross-institutional diagnostic performance.
    \item Selecting $X_{\text{test}} = X_1$, we test the model's generalization ability on different system data from the same institution.
    \item Selecting $X_2, X_3, X_4, X_5$, and $X_6$, respectively, and combining them with the corresponding system-specific data from $X_1$, we create five new datasets. These new datasets are utilized to evaluate the model's generalization ability across different systems and institutions.
\end{itemize}

Our goal is to enable models to possess cross-institution and multi-system diagnostic capabilities. The fundamental challenge faced in mathematical analysis is how to leverage latent knowledge to constrain the model's freedom within a vast space of heterogeneous data samples, thereby enabling the model to achieve an approximate global optimal solution in Eq.\eqref{eq1}. In our experiments, the effective fine-tuning techniques are exploited to restrict the pre-trained model's degrees of freedom, resulting in an improved generalization performance in the feature space.

\subsection{Framework Overview}

This section provides a comprehensive introduction to the method presented in this article, with applicability to large-scale, multi-institution, and multi-system in the field of radiology, as illustrated in Figure \ref{fig:overflow}. The method is based on a dataset comprising 332673 radiology records obtained from multi-institution and multi-system sources and designed to address the clinical challenge of radiology reports originating from multiple institutions. Here, we present an overall architecture of the proposed method by leveraging medical knowledge in the field of radiology, followed by a detailed description of each phase. The proposed method comprises five distinct phases: data acquisition, data preprocessing, abstract feature extraction, radiology report generation, and experts evaluation.

\paragraph{1) \textit{Phase 1 (Data Acquisition):}} As an extremely important department in hospitals or medical institutions, the radiology department has different writing habits and styles of radiology reports, making it difficult to unify. Acknowledging this distinctive nature of report data within the radiology department, it is critical to address the inherent challenges in obtaining such data. This inherent challenge presents formidable obstacles when seeking to employ LLMs in the field of radiology. Therefore, in collaboration with Xiangya Second Hospital, we acquire radiology reports from six distinct institutions and five disparate systems. This meticulous data acquisition effort ensures the availability of high-quality data requisite for the training of our expected LLM, ChatRadio-Valuer. The description of the data acquisition can be found in Section \textbf{\ref{sec:Collaboration with the Second Xiangya Hospital of China}} and \textbf{\ref{sec:Data Description}}.

\paragraph{2) \textit{Phase 2 (Data Preprocessing):}} LLMs necessitate the use of high-quality data for their pre-training, a critical property that distinguishes them from smaller-scale language models. The model capacity of LLMs is profoundly influenced by the pre-training corpus and its associated preprocessing methodologies, as expounded in \cite{zhao2023survey}. The presence of a substantial number of bad samples within the corpus can significantly degrade the performance of the model. To alleviate the detrimental impact of such bad samples on model performance and enhance models' adaptability and generalizability, we collaborate closely with domain experts from hospitals to devise a method for the data preprocessing of these bad samples, which is discussed in Section \textbf{\ref{sec:Data Preprocessing}}.

\paragraph{3) \textit{Phase 3 (Abstract Feature Extraction):}} LLMs invariably demand a heavy investment of human resources, computational resources, financial cost, and time consumption, presenting formidable challenges to researchers. Consequently, our method, following exhaustive investigations and rigorous experimental validation, leverages the Llama2 series of models. Additionally, we have employed proprietary datasets, acquired and preprocessed in the previous two phases, for enhancing the performance of ChatRadio-Valuer. Comprehensive insights into this abstract feature extraction process are detailed in Section \textbf{\ref{sec:How to Fine-tune ChatRadio-Valuer?}}.

\paragraph{4) \textit{Phase 4 (Radiology Reports Generation):}} Leveraging the power of our proposed ChatRadio-Valuer, we can generate causal inferences with a specific set of configurations. The provided "Finding" is utilized as the input to the LLM, from which the "Impression", the output of the LLM, is derived. The explicit process to implement ChatRadio-Valuer for generating radiology reports refers to the Section \textbf{\ref{sec:How to Inference ChatRadio-Valuer?}}.

\paragraph{5) \textit{Phase 5 (Experts Evaluation):}} Currently, there still exists controversy over the evaluation indicators for LLMs in the radiology field. Traditional similarity indicators fall short in accurately reflecting the specific performance of the model in this field, let alone measuring their practical applicability. In response to this challenge, our method, in collaboration with domain experts from medical institutions, introduces an evaluation framework tailored for medical LLMs. This framework not only excels in conventional similarity metrics as engineering ones but also offers a comprehensive assessment of the models' practical clinical utility. Comparative analysis against the state-of-the-art (SOTA) methods reveals that the proposed ChatRadio-Valuer maintains a competitive advantage. A detailed exposition of this evaluation methodology can be found in Section \textbf{\ref{sec:PerformanceEvaluation}}.

\subsection{How to Fine-tune ChatRadio-Valuer?}\label{sec:How to Fine-tune ChatRadio-Valuer?}

\subsubsection{Architecture of the Llama Model}
\begin{figure}[htbp]
    \centering
    \includegraphics[width=\textwidth]{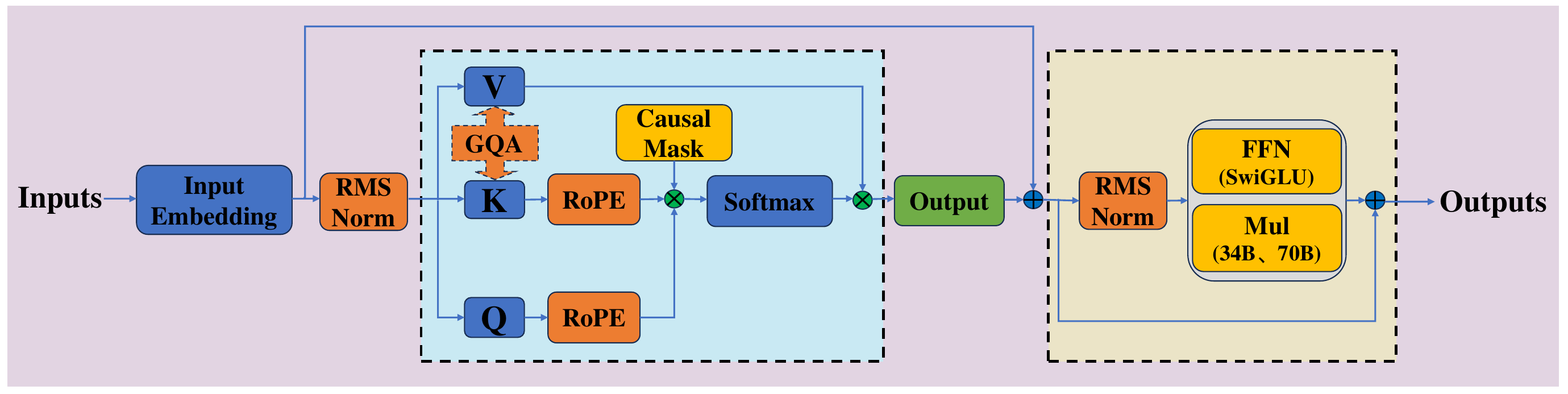}
    \caption{The architecture diagram of Llama 2. The model structure of Llama 2 is basically consistent with the standard Transformer Decoder structure, mainly composed of 32 Transformer Blocks}
    \label{fig:main2}
\end{figure}

In artificial intelligence, Llama has won the affection of a vast user base with its unique charm and unparalleled intelligence. However, this is not the end of the story. Llama2 not only inherits all the advantages of its predecessors but also brings significant innovations and improvements in many aspects. Taking into account the special problems of diagnosis of a variety of diseases based on radiology reports, the considerations for the foundation model are detailed as follows:

\begin{enumerate}
    \item \textbf{Large-scale Dataset:} The training dataset of the Llama2 model is extremely extensive. Building upon Llama, Llama2 expanded its training corpus by approximately 40\%, comprising a staggering 2 trillion tokens. This vast dataset equips the model with the ability to learn from a diverse range of data sources, empowering it to handle heterogeneous and diverse data effectively.
    \item \textbf{Expansion of Context Length:} Llama2 extend the context length from Llama's 2048 tokens to an impressive 4096 tokens. This expansion signifies that Llama2 is better equipped to comprehend and handle lengthier text passages. When dealing with heterogeneous data, there may be a greater amount of textual information to consider, and the extended context length aids in capturing this information more comprehensively.
    \item \textbf{Multi-task Adaptability:} Llama2 exhibits multi-task adaptability, indicating its capability to concurrently perform multiple tasks when dealing with heterogeneous data. This enables it to more comprehensively leverage the diversity and complexity of the data.
    \item \textbf{Model Architecture Improvement:} Llama2 inherits most of Llama's pre-training settings and model architecture \cite{touvron2023llama1} but also introduces several enhancements, including expanding matrix dimensions in the Feedforward Neural Network (FFN) module and introducing Grouped Query Attention (GQA) \cite{touvron2023llama2}. These improvements can enhance the model's performance and efficiency, making it better suited to heterogeneous data. A schematic diagram of Llama2 is presented in Figure \ref{fig:main2}.
\end{enumerate}

Some technical details are described as follows:

\begin{itemize}
    \item In the model's Norm layer, RMSNorm is used without re-centering operations (the removal of the mean term), significantly improving model speed \cite{zhang2019root}. The RMSNorm formula is as follows:
    \begin{equation} \label{eq2}
        \text{RMSNorm}(x) = \frac{x}{\sqrt{\text{Mean}(x^2) + \epsilon}}
    \end{equation}
    where $x$ represents the input vector or tensor, typically the output of a neural network layer; $\epsilon$ is a small positive constant, usually used to prevent the denominator from being zero, ensuring numerical stability. It is typically a very small value, such as $1.0e^{-5}$ or $1.0e^{-6}$.
    
    \item SwiGLU activation function is employed in the Feed-Forward Network (FFN) to enhance performance. The implementation of the SwiGLU activation function is as follows:
    \begin{equation} \label{eq3}
        SwiGLU(x, W, V, b, c, \beta) = \text{Swish}_\beta(xW + b) \odot (xV + c)
    \end{equation}
    where $Swish$ is defined as \(Swish = x \cdot \sigma(\beta x)\); $GLU$ is represented as \(GLU(x) = \sigma(Wx + b) \odot (Vx + c)\); $W$ and $V$ denote weight matrices; $b$ and $c$ represent biases; The symbol $\odot$ represents the element-wise Hadamard product, also known as element-wise multiplication.
    \item RoPE (Rotary Position Embedding) is a technique used in neural networks to represent sequential position information, particularly in natural language processing tasks. The core idea behind RoPE is to introduce a rotational operation to represent position information, avoiding the fixed nature of traditional positional embeddings. Specifically, the formula for RoPE's positional embedding is as follows:
    \begin{equation} \label{eq4}
        RoPE(\text{position}, d) = R_d \cdot P_{\text{position}}
    \end{equation}
    where $RoPE(position, d)$ represents the RoPE position embedding at a specific position; $d$ denotes the embedding dimension; $R_d$ denotes a rotation matrix with dimensions of $d$, utilized for rotating the position embedding; $P_{position}$ denotes an original position embedding vector, typically generated by sine and cosine functions. However, in RoPE, it is not static but can be rotated to adapt to different tasks.
    \item GQA is the abbreviation for grouped-query attention, which is a variant of the attention mechanism. In GQA, query heads are divided into multiple groups, each of which shares a single key and value matrix. This mechanism serves to reduce the size of the key-value cache during inference and significantly enhances inference throughput.
\end{itemize}

Based on the selected Llama2 family (i.e., Llama2, Llama2-Chat, and Llama2-Chinese-Chat), the model fine-tuning process in this paper is described in Algorithm \ref{alg:train} and outlined as follows:

\begin{algorithm}[H]
\caption{Train ChatRadio-Valuer.}\label{alg:train}
\begin{algorithmic}[1]
\Require 
Pre-trained model $\boldsymbol{f}_{0}$; 
Quantization parameter $\boldsymbol{\theta}_{quantization}$; 
Training parameters $\boldsymbol{\theta}_{tr}$; 
PEFT (LoRA) parameters $\boldsymbol{\theta}_{peft-lora}$; 
Training prompts $\boldsymbol{X}^{tr}_{prompt}=\left\{\boldsymbol{sp}^{tr}_{1}, \boldsymbol{sp}^{tr}_{2}, \ldots, \boldsymbol{sp}^{tr}_{5}\right\}$; Training labels $\boldsymbol{L}_{tr}=\left\{\boldsymbol{y}_{tr1},\boldsymbol{y}_{tr2},\ldots,\boldsymbol{y}_{trj}\right\}$
\Ensure 
Best prompt candidate index $\boldsymbol{idx}$;
ChatRadio-Valuer model $\boldsymbol{f}$ and parameters $\boldsymbol{\theta}$
\State$\boldsymbol{f}_{init} = InitQuantizedModel(\boldsymbol{f}_{0}, \boldsymbol{\theta}_{quantization})$
\State $\boldsymbol{\theta}_{config}=ModeLoRA(LoadTrConfig(\boldsymbol{\theta}_{tr}),\boldsymbol{\theta}_{peft-lora})$
\For{$\mathbf{idxPrompt} := \textbf{from} \textbf{ 1} \textbf{ to} \textbf{ i}$} 
\For{$\mathbf{TrEpoch} := \textbf{from} \textbf{ 1} \textbf{ to} \textbf{ SmallEpoch}$}
\State $\boldsymbol{f}_{idxPrompt} = TrainSFTModel(\boldsymbol{f}, \boldsymbol{sp}^{tr}_{idxPrompt}, \boldsymbol{\theta}_{config})$
\State $\boldsymbol{\theta}_{idxPrompt} = UpdateParameter(\left\{\boldsymbol{y}_{tr1},\boldsymbol{y}_{tr2},\ldots,\boldsymbol{y}_{trj}\right\})$\
\EndFor
\EndFor
\State $\boldsymbol{idx} = FindBestPrompt((\boldsymbol{f}_{1},\boldsymbol{\theta}_{1}), (\boldsymbol{f}_{2},\boldsymbol{\theta}_{2}),\ldots, (\boldsymbol{f}_{i},\boldsymbol{\theta}_{i}))$
\For{$\mathbf{TrEpoch} := \textbf{from} \textbf{ SmallEpoch} \textbf{ to} \textbf{ FullEpoch}$} 
\State $\boldsymbol{f} = ContinueTrainSFTModel(\boldsymbol{f}_{idx}, \boldsymbol{sp}^{tr}_{idx}, \boldsymbol{\theta}_{idx})$
\State $\boldsymbol{\theta} = ContinueUpdateParameter(\left\{\boldsymbol{y}_{tr1},\boldsymbol{y}_{tr2},\ldots,\boldsymbol{y}_{trj}\right\})$\
\EndFor
\State \Return $\boldsymbol{idx}, \boldsymbol{f}, \boldsymbol{\theta}$
\end{algorithmic}
\end{algorithm}

\begin{enumerate}
    \item \textbf{Model Quantization:} Leveraging the Transformers' model quantization library, we employ the BitsAndBytesConfig interface for model quantization. Essentially, this process involves converting model weights into int4 format through quantization layers and placing them on the GPU. Referring to Algorithm \ref{alg:train}, we initialize the pre-train model $\boldsymbol{f}_{0}$ with the int4 quantization configuration $\boldsymbol{\theta}_{quantization}$ to prepare for further model preparation. The core computation is carried out on the CUDA, reducing memory consumption and improving efficiency.
    \item \textbf{Model Fine-tuning:} We incorporate LoRA (Low-Rank Adaptation) weights $\boldsymbol{\theta}_{peft-lora}$ and training parameters $\boldsymbol{\theta}_{tr}$ shown in Algorithm \ref{alg:train}. In the first step, we fix the parameters of the model's transformer component, focusing solely on training embeddings. This is done to adapt the newly added small-sample vectors without significantly interfering with the original model. Additionally, supervised Prompt data is utilized for fine-tuning, aiding in the selection of the most suitable prompts. In the second step, training continues on the remaining majority of samples while simultaneously updating LoRA parameters. The LoRA weights are then merged back into the initialized model $\boldsymbol{f}_{init}$, resulting in fine-tuned model $f$ with model parameters $\theta$.
    \item \textbf{Efficient Prompt Selection:} During the stage of fine-tuning, to effectively activate the generalizability and semantic capability of LLMs, prompt selection plays a significant role. However, fine-tuning any LLMs thoroughly is a time-consuming project, which requires sufficient computational resources and time capacity. Therefore, we employ small-epoch fine-tuning with our five expert-curated training prompts $\boldsymbol{X}^{tr}_{prompt}=\left\{\boldsymbol{sp}^{tr}_{1}, \boldsymbol{sp}^{tr}_{2}, \ldots, \boldsymbol{sp}^{tr}_{5}\right\}$ for time efficiency. We find the best prompt and its index $\boldsymbol{idx}$ through this small-epoch fine-tuning, and then inherit the model and parameters pair $(\boldsymbol{f}_{\boldsymbol{idx}},\boldsymbol{\theta}_{\boldsymbol{idx}})$ to continue the complete model fine-tuning mention above.
\end{enumerate}

\subsubsection{Prompt Generation}

With the meticulously processed dataset proposed in Section \ref{sec:Data Preprocessing}, our objective is to generate complete prompts, subsequently employed by LLMs to yield effective causal inferences, contributing to either training or evaluation phases. Before introducing the proposed prompt, we first explicate the parts of the structure, which is accessible via a string and divided into three components: system description, instruction and input. The system description component is typically invoked at the outset to demarcate the task and constrain the instruction's behavior, while the instruction component serves to provide direction for radiology report generation and the input component contains clinical radiology reports. The model output is generated by the response component, which serves as the basis for our dynamic prompt and iterative optimization framework.

\begin{figure}[htbp!]
    \centering
    \includegraphics[width=0.8\textwidth]{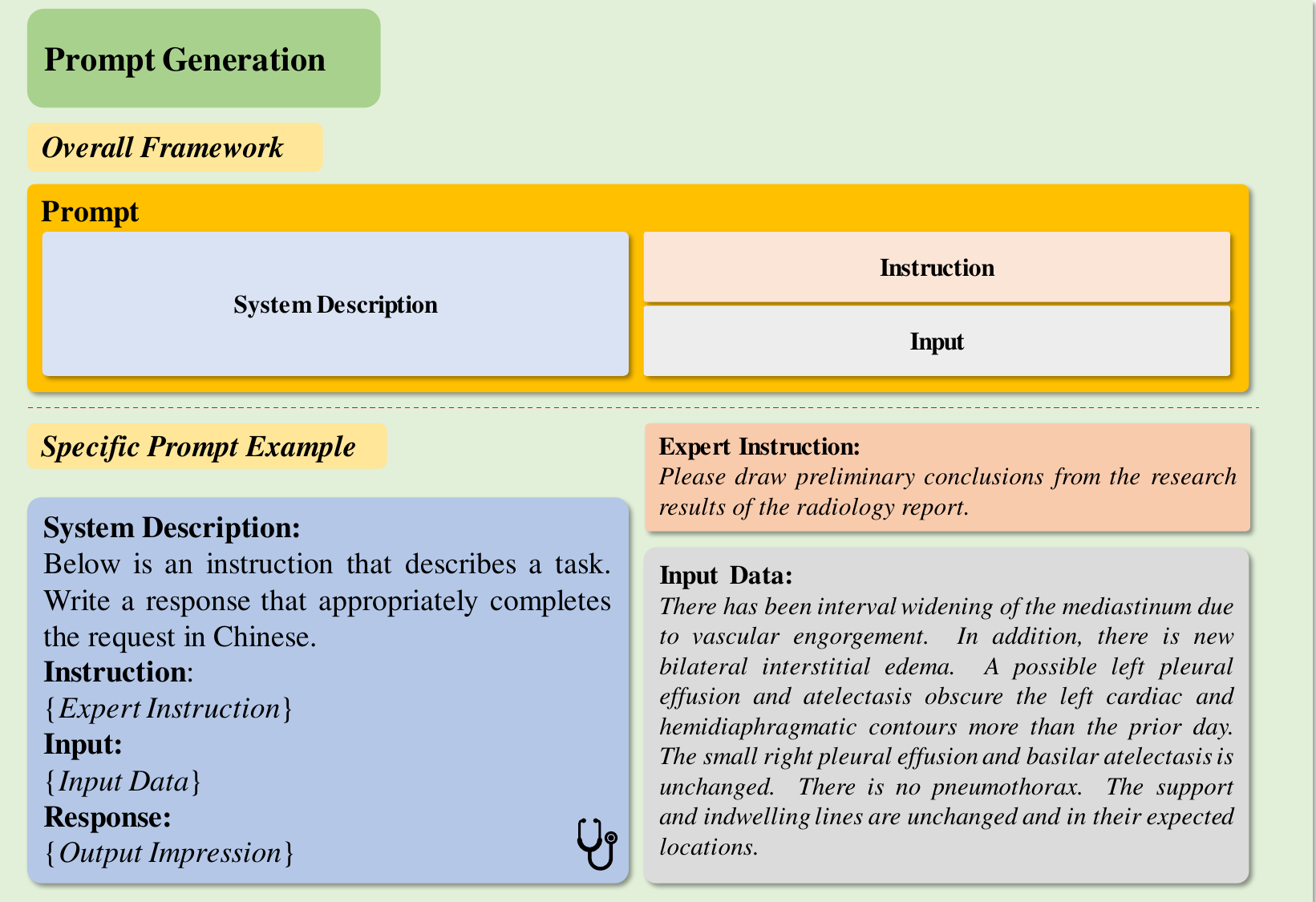}
    \caption{Prompt generation overview. The overall framework contains three parts, system description, instruction, and input, which collaboratively constitute a prompt. Within a prompt example (purple), expert instruction and input data on its right are inserted to the \{\textit{Expert Instruction}\} and \{\textit{Input Data}\}, respectively. The derived impression is in the \{\textit{Output Impression}\}.}
    \label{fig:promptgeneration}
\end{figure}

Prior studies have used fixed-form prompts for straightforward tasks that could be easily generalized. However, these prompts lack the necessary prior knowledge for complex tasks and domain-specific datasets, resulting in low performance \cite{ma2023impressiongpt}. Thus, we propose a hypothesis that constructs dynamic prompts from relevant domain-specific corpora and can thus enhance the model's comprehension and perception.

Specifically, in addition to the corpus containing preprocessed multi-institution and multi-system data, domain experts contribute to the formulation of five prompt templates, instructions to LLMs, tailored to this study. Their professional advice helps provide refined instructions to LLMs to activate their ability to adapt to radiology domain and generate meaningful domain-specified results. Definitely referring to Algorithm \ref{alg:eval}, we combine the synthesized expert-curated instruction and system description set $\boldsymbol{X}_{pt}$ with the input data. As an illustrative example, one of these templates is shown in Figure \ref{fig:promptgeneration}, where \textit{\textbf{Input Data}} is for data insertion and \textit{\textbf{Expert Instruction}} is for instruction insertion. The complete prompt generation scheme is illustrated in Figure \ref{fig:promptgeneration}. Following this approach, five distinct prompt sets are generated, fully prepared for subsequent training and evaluation stages. 

\algdef{SE}[DOWHILE]{Do}{doWhile}{\algorithmicdo}[1]{\algorithmicwhile\ #1}%
\begin{algorithm}[H]
\caption{Inference $\&$ Evaluate Reports.}\label{alg:eval}
\begin{algorithmic}[1]
\Require
Pre-trained model $\boldsymbol{f}_{0}$; 
Expert-curated prompt templates $\boldsymbol{X}_{pt}=\left\{\boldsymbol{t}_{1},\boldsymbol{t}_{2},\ldots,\boldsymbol{t}_{i}\right\}$; 
Best prompt index $\boldsymbol{idx}$; 
Test-institution data $\boldsymbol{X}_{test}=\left\{(\boldsymbol{x}^{c}_{1}, \boldsymbol{y}^{c}_{1}),(\boldsymbol{x}^{c}_{2}, \boldsymbol{y}^{c}_{2}),\ldots,(\boldsymbol{x}^{c}_{l}, \boldsymbol{y}^{c}_{l})\right\}$; 
Evaluation prompts $\boldsymbol{X}^{eval}_{prompt}=\left\{\boldsymbol{sp}^{eval}_{1},\boldsymbol{sp}^{eval}_{2},\ldots,\boldsymbol{sp}^{eval}_{i}\right\}$;
Inference Configuration $\boldsymbol{\gamma}$;
Evaluation labels $\boldsymbol{L}_{eval}=\left\{\boldsymbol{y}_{e1},\boldsymbol{y}_{e2},\ldots,\boldsymbol{y}_{ej}\right\}$; 
ChatRadio-Valuer parameters $\boldsymbol{\theta}$;
\Ensure
Expert-checked inference reports $\boldsymbol{I}$; 
Rouge scores $\boldsymbol{R}$; 
Expert-scored result report $\boldsymbol{C}$
\State $\boldsymbol{f} = InitModel(\boldsymbol{f}_{0})$
\State $\boldsymbol{f}_{eval} = LoadParameter(\boldsymbol{f}, \boldsymbol{\theta})$
\State$\boldsymbol{sp}_{c}=SynthesizePrompt(\boldsymbol{t}_{\boldsymbol{idx}}, \left\{\boldsymbol{x}^{c}_{1}, \boldsymbol{x}^{c}_{2}, \ldots, \boldsymbol{x}^{c}_{l}\right\})$
\Do
\State $\boldsymbol{I}_{eval} = ChineseTextSegmentation(\boldsymbol{f}_{eval}(\boldsymbol{sp}^{eval}_{\boldsymbol{idx}}), \boldsymbol{\gamma})$
\State $\boldsymbol{I}_{c} = ChineseTextSegmentation(\boldsymbol{f}_{eval}(\boldsymbol{sp}_{c}), \boldsymbol{\gamma})$
\State $\boldsymbol{I} = \{\boldsymbol{I}_{eval}, \boldsymbol{I}_{c2}\} $
\doWhile $ NANCheck(\boldsymbol{I})$ $ \&$ $ TimeOutCheck(\boldsymbol{I})$
\State $\boldsymbol{R}_{eval} = RougeTest(\boldsymbol{I}_{eval}, \left\{\boldsymbol{y}_{e1},\boldsymbol{y}_{e2},\ldots,\boldsymbol{y}_{ej}\right\}) $
\State $\boldsymbol{R}_{c} = RougeTest(\boldsymbol{I}_{c}, \left\{\boldsymbol{y}^{c}_{1},\boldsymbol{y}^{c}_{2},\ldots,\boldsymbol{y}^{c}_{l}\right\}) $
\State $\boldsymbol{R} = \{\boldsymbol{R}_{eval}, \boldsymbol{R}_{c}\} $
\State $\boldsymbol{C} = ExpertScoring(\boldsymbol{I}, \boldsymbol{R})$
\State \Return $\boldsymbol{I}, \boldsymbol{R}, \boldsymbol{C}$
\end{algorithmic}
\end{algorithm}

\subsection{How to Inference ChatRadio-Valuer?}\label{sec:How to Inference ChatRadio-Valuer?}
In the final phase, model reasoning testing and evaluation, clinging to the power of LLMs, including our proposed ChatRadio-Valuer especially, we exert them to generate casual inferences on the evaluation dataset $\boldsymbol{I}$  for evaluating ChatRadio-Valuer's performance over the other selected LLMs'. 

In Algorithm \ref{alg:eval}, we have decided the best evaluation prompt template candidate $\boldsymbol{t_{idx}}$ with its index $\boldsymbol{idx}$. This template is also implemented in this module for saving computation resources and better exerting the capability of LLMs (i.e., no more prompt template selection is required due to its reliable credits in previous performance comparisons). Accordingly, this template indicates the expected evaluation prompts $\boldsymbol{sp}^{eval}_{\boldsymbol{idx}}$. In terms of the test-institution prompt sets, following the same data preprocessing schematic in the second stage, we synthesize the cleaned test-institution data $\boldsymbol{X}_{test}$ and the designated template candidate $\boldsymbol{t_{idx}}$ into the final test-institution prompts $\boldsymbol{sp}_{c}$. 

With the massive preparation work of multi-institution and multi-system prompts finished, we carefully construct our LLM pool for the final evaluation, containing total 15 LLMs, including our ChatRadio-Valuer. For each LLM, weights are accordingly initialized (i.e., ChatRadio-Valuer is initiated with SFT weights but others are with their official ones) and prepared for causal inference. With all inference configuration $\boldsymbol{\gamma}$, including \textit{top-k}, \textit{top-p}, \textit{temperature}, etc., are specified, we leverage the power of these LLMs to inference casually, obtaining the reports $\boldsymbol{I}_{eval}$ and $\boldsymbol{I}_{c}$. During the generation process, we pay great attention to the quality of the generated results. However, limited by the highly hardware-dependent properties of LLMs and randomness within their outputs, results sometimes encounter problems including the null output and repetitive words, leading to inaccurate representation of LLMs' performances on generating casual inferences, specifically the quality of the reports without meaningless values. In order to solve these problems, we incorporate two modules to check whether the null output is generated and repetitive words are obtained during generation. These two modules maintain the quality of the final results in avoid of interference caused by meaningless values. In addition, more language-specifically, we leverage Chinese text segmentation to the results to improve the accuracy of ROUGE evaluation towards Chinese contexts. This segmentation provides a valuable vision for bilingual or multi-lingual tasks.

\section{Performance Evaluation}\label{sec:PerformanceEvaluation}

\subsection{Performance Evaluation Indices}\label{sec:Performance Evaluation Indexes}
To confirm the effectiveness, generalization, and transferability of our proposed ChatRadio-Valuer, we conduct a comprehensive analysis using engineering indexes and clinical evaluation from the aspects of feasibility, clinical performance, and application cost.
 
For feasibility, we evaluate the performance of the models by the widely-used engineering index ROUGE \cite{lin2004rouge} and report the $F_1$ scores for ROUGE-N and ROUGE-L (denoted as R-1, R-2, and R-L), which measure the word-level N-gram-overlap and longest common sequence between the reference summaries and the candidate summaries respectively, as shown from Eq.\eqref{eq5}-Eq.\eqref{eq8}. $P$ is denoted as the percentage of defects correctly assigned, while $R$ represents the ratio of accurately detected defects to total true defects. $F_1$ measures the trade-off between Precision($P$) and Recall($R$).

 \begin{equation} \label{eq5}
ROUGE-N=\frac{\sum_{S\in\lbrace ReferenceSummaries}{\sum_{{gram}_n\in S}{Count_{match}(gram_n)}}}{\sum_{S\in\lbrace ReferenceSummaries}{\sum_{{gram}_n\in S}{Count(gram_n)}}}
\end{equation}

 \begin{equation} \label{eq6}
R=\frac{LCS(ReferenceSummaries,CandidateSummaries)}{m}
\end{equation}

 \begin{equation} \label{eq7}
P=\frac{LCS(ReferenceSummaries,CandidateSummaries)}{n}
\end{equation}

 \begin{equation} \label{eq8}
F_1=\frac{2RP}{R+P}
\end{equation}

where $n$ stands for the length of the n-gram, $gram_n$; 
$Count_{match}(gram_n)$ is the maximum number of the n-grams co-occurring in a candidate summary and a set of reference summaries. \textit{LCS} is the length of the longest common subsequence of Reference-Summaries and Candidate-Summaries. $m$ denotes the length of Reference-Summaries and $n$ is the length of Candidate-Summaries.

For clinical performance, we assess ChatRadio-Valuer using an array of clinical indexes to test its performance, as illustrated in Figure \ref{fig5}. For each index, we utilize a 100-point scale and divide it into quintiles every 20 points. These clinical indexes include "Understandability" which denotes the model's capacity to be comprehended and interpreted by clinicians and relevant physicians. "Coherence" means the model's ability to maintain logical consistency and unified structure in its outputs. Furthermore, "Relevance" means the model's capacity to generate related information and insights for the current clinical context. "Conciseness" represents the importance of the model's output being concise and reducing redundant information, ensuring that it effectively conveys essential clinical knowledge. "Clinical Utility" emphasizes the practical value of the model's outputs, measuring its capacity to inform and enhance clinical decision-making processes. Moreover, the evaluation also examines the potential for "Missed Diagnosis" emphasizing the model's ability to minimize instances where clinically significant conditions are overlooked or underemphasized. On the contrary, "Overdiagnosis" assesses the condition where the model may lead to excessive diagnoses or overutilization of medical interventions. These clinical indexes inform the assessment of the model's effectiveness, ensuring its effectiveness and appropriateness in the clinical domain. In addition to the basic interpretation of these metrics, the most essential factors for clinical utilities and significance are "Missed Diagnosis" and "Overdiagnosis", explained more thoroughly as follows: 

\paragraph{Missed Diagnosis} Missed diagnosis, or false negative in medical diagnostics, occurs when healthcare practitioners fail to detect a medical condition despite its actual presence. This diagnostic error stems from the inability of clinical assessments, tests, or screenings to accurately identify the condition, resulting in a lack of appropriate treatment. The consequences of missed diagnosis range from delayed therapy initiation to disease progression, with contributing factors including diagnostic modality limitations, inconspicuous symptoms, and cognitive biases. Timely and accurate diagnosis is fundamental to effective healthcare. Therefore, it is necessary to reduce the occurrence of missed diagnosis for improved patient outcomes and healthcare quality.

\paragraph{Overdiagnosis} Overdiagnosis is characterized by the erroneous or unnecessary identification of a medical condition upon closer examination. This diagnostic error arises when a condition is diagnosed but would not have caused harm or clinical symptoms during the patient's lifetime. Overdiagnosis often leads to unnecessary medical interventions, exposing patients to potential risks without significant benefits. It is particularly relevant in conditions with broad diagnostic criteria, where the boundaries between normal and pathological states are unclear. For example, early detection of certain malignancies can result in overdiagnosis if the cancer has remained asymptomatic throughout the patient's lifetime. Recognizing and addressing overdiagnosis are critical aspects of modern healthcare, influencing resource allocation, patient well-being, and healthcare system sustainability.

As for application metrics, the assessment of practicality is critical for ChatRadio-Valuer. In this context, two critical indexes are "Time Cost" (i.e. "Fine-Tuning Time" and "Testing Time") and "Parameter Count". "Fine-tuning time" means the temporal duration during the iterative training process where the model acquires its knowledge and optimization. The duration of the training phase is of great importance, as it directly impacts the time-to-model readiness and the overall efficiency of the development workflow. "Testing Time"  is an essential measure that gauges the computational expenses incurred during the model's execution, including both the temporal and computational resources required for real-time deployment. This metric carries great significance, particularly in resource-limited settings where efficient use of computational resources is vital. "Parameter Count" is another vital evaluation metric, indicating the model's demand for memory storage during execution. This aspect is of great importance, especially when dealing with data-intensive applications. An effective evaluation of memory size ensures that the model remains deployable on hardware configurations with adequate memory resources while avoiding performance degradation. In a word, these indexes serve as crucial benchmarks in the assessment of the model, allowing for a comprehensive evaluation of its computational efficiency and resource utilization in the specific operational setting.

\subsection{Experiment Setup}

We initially preprocess the multi-system data from multiple institutions (shown in Section \ref{sec:Data Preprocessing}) and established datasets for training and evaluation by combining these data with five prompt templates. After data preprocessing, we implement supervised fine-tuning (SFT) on ChatRadio-Valuer using these cleaned and standardized data. We also employ evaluation on the SFT ChatRadio-Valuer. To achieve reduced computational resource requirements, faster adaptation, and enhanced accessibility, we apply low-rank adaptation (LoRA), a technique for parameter-efficient fine-tuning (PEFT), and 4-bit quantization during the SFT stage. Technically, LoRA enhances the efficiency of adapting large, pre-trained language models to specific tasks by introducing trainable rank decomposition matrices. It significantly reduces the number of trainable parameters while maintaining high model quality when compared to traditional fine-tuning. In terms of quantization, it simplifies numerical representation by reducing data precision. In LLMs, it involves the conversion of high-precision floating-point values into lower-precision fixed-point representations, effectively diminishing memory and computational requirements. Quantized LLMs offer enhanced resource efficiency and compatibility with a wide array of hardware platforms. In the context of ChatRadio-Valuer, we implement 4-bit precision quantization. 

Before the training stage, we investigate the well-known LLMs from medical perspective and constructed LLM pool for model selection. We mainly focus on 6 aspects, indicated in Figure \ref{fig:modelpool}, domain adaptability, compatibility with medical standards, bilingual, open source, parameter efficiency, and cost and licensing. A brief introduction to these metrics is listed below:

\begin{itemize}
    \item \textbf{Domain Adaptability:} Analyze the LLM's amenability to fine-tuning and adaptation within specific medical domains or subfields, such as radiology and more detailed applications within this field.
    \item \textbf{Compatibility with Medical Standards:} Evaluate the LLM's adherence to established medical standards and terminologies to facilitate seamless integration with healthcare systems.
    \item \textbf{Bilingual:} Ascertain the LLM's proficiency in effectively processing and understanding Chinese and English at the same time, which holds significance for medical data and research encompassing diverse linguistic regions.
    \item \textbf{Open Source:} Investigate whether the LLM is released under an open-source licensing model, thereby fostering transparency, customizable functionality, and the potential for collaborative contributions from the broader community. Open-source models often garner wider adoption and robust support.
    \item \textbf{Parameter Efficiency:} Deliberate upon the model's computational efficiency, encompassing considerations of inference time, memory utilization, and resource demands. Efficiency assumes paramount importance in applications demanding real-time processing or operating within resource-constrained and time-constrained environments
    \item \textbf{Cost and Licensing:} Conduct a comprehensive cost analysis pertaining to the utilization of the LLM, inclusive of any licensing fees or usage constraints, while ensuring alignment with budgetary constraints and adherence to legal stipulations.
\end{itemize}

Eventually, we choose 12 SOTA LLMs and take 4 of them, Llama2-7B, Llama2-Chat-7B, Llama2-Chinese-Chat-7B and ChatGLM2-6B for fine-tuning. During the training stage, we conduct a total of three training epochs. The initial learning rate is set to $1.41e^{-5}$, with a steady decrease as training steps progressed. The batch size is fixed at 64, and the maximum input/output sequence length are set to 512. To alleviate GPU memory load, we use gradient accumulation steps, which are set to 16. In terms of LoRA, we assign values of 64 and 16 to the parameters $r$ and $\alpha$, respectively. During the evaluation stage, all inferences share the same configuration, with a maximum token generation length of 512. Temperature, top-k, and top-p values are set to 1.0, 50, and 1.0, respectively. The prevailing algorithms as the baseline to construct a LLM pool are shown in Figure \ref{fig:modelpool} and explained as follows:

\begin{figure}[htbp!]
    \centering
    \includegraphics[width=1\textwidth]{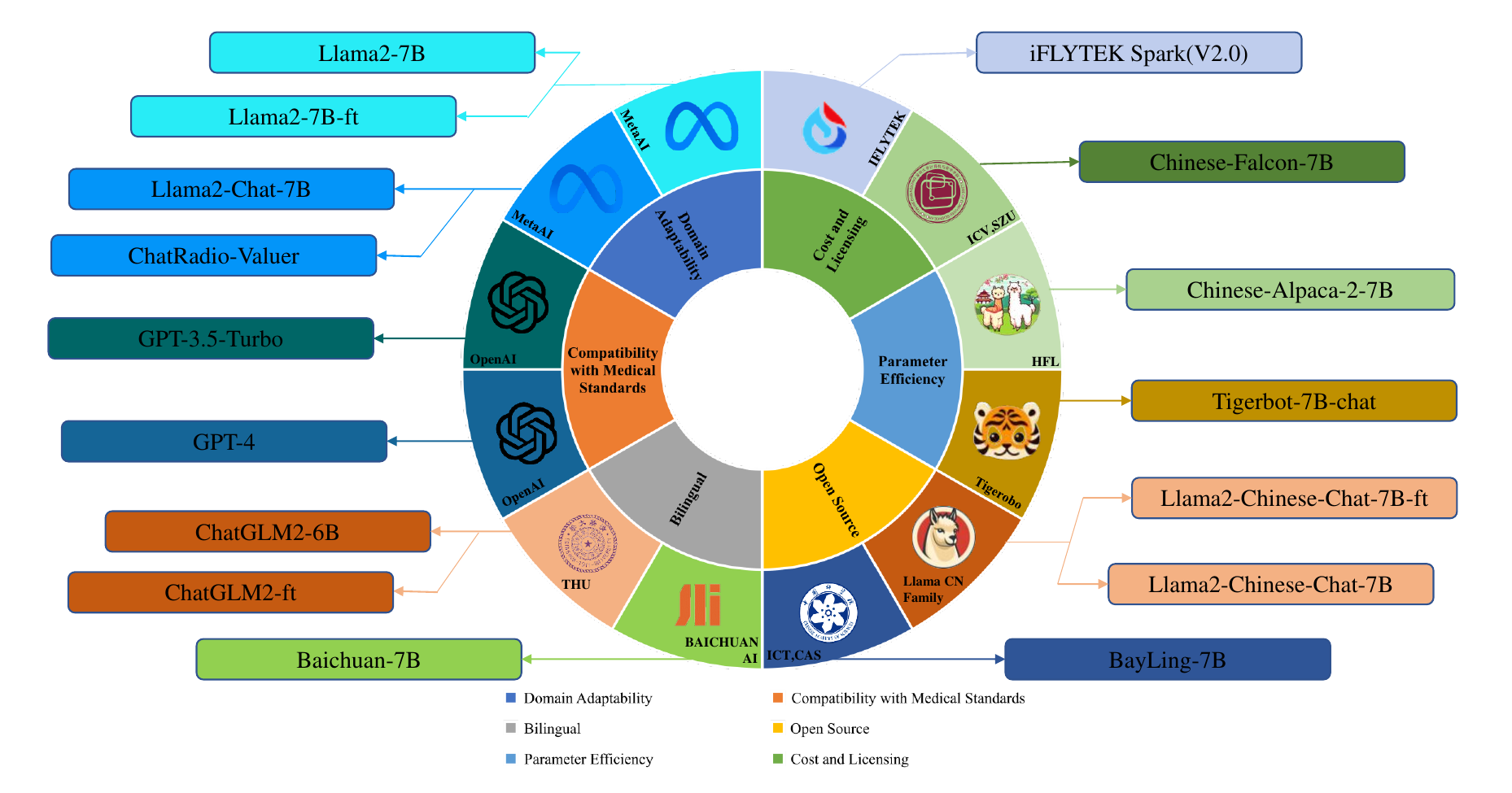}
    \caption{LLM pool for model selection. Considering the scene of medical application, six aspects are considered: domain adaptability, compatibility with medical standards, bilingual, open source, parameter efficiency, and cost and licensing. 15 SOTA LLMs (12 baseline models, 3 fine-tuning pairs, and 1 fine-tuned ChatGLM2-6B) from 10 organizations jointly are established by the LLM pool.}
    \label{fig:modelpool}
\end{figure}

\textbf{Baseline Methods: }

\textbf{1) Llama2:} Llama2 \cite{touvron2023llama} is a collection that comprises a meticulously curated set of pre-trained generative text models that have been expertly fine-tuned. These models operate as auto-regressive language models and are founded upon an optimized transformer architecture. The fine-tuned variants incorporate SFT and reinforcement learning with input from human feedback (RLHF) \cite{knox2011augmenting} to align with human preferences regarding both utility and safety. Our model selection includes Llama2-7B and Llama2-Chat-7B.

\textbf{2) Llama2-Chinese:} Llama2-Chinese \cite{githubGitHubFlagAlphaLlama2Chinese} is a collection containing LLMs pre-trained on a large-scale Chinese corpus, leveraging high-quality Chinese language datasets totaling 200 billion characters for fine-tuning. While this approach is costly, requiring significant resources in terms of both high-quality Chinese data and computational resources, its great advantage lies in enhancing Chinese language capabilities at the model's core, achieving fundamental improvement and imparting strong Chinese language proficiency to the large model. In our study, we carefully select Llama2-Chinese-Chat-7B as the baseline method.

\textbf{3) GPT-3.5-Turbo/GPT-4:} ChatGPT and GPT4, both developed by OpenAI, are influential large language models. ChatGPT, also known as GPT-3.5-turbo, is an advancement based on GPT-2 and GPT-3, with its training process heavily influenced by instructGPT  \cite{ouyang2022training}. A key distinction from GPT-3 \cite{brown2020language} is the incorporation of RLHF \cite{knox2011augmenting}, which refines model output through human feedback. This approach enhances the model's ability to rank results effectively. ChatGPT excels in language comprehension, accommodating diverse expressions and queries. Its extensive knowledge base answers frequently asked questions and provides valuable information. GPT-4, being the successor of GPT-3, may possess enhanced capabilities. In our experiments, both ChatGPT and GPT4 are utilized.

\textbf{4) ChatGLM2-6B:} ChatGLM2-6B \cite{zeng2022glm, du2022glm} is the second-generation bilingual chat model based on the open-source ChatGLM-6B framework. It has undergone pre-training with 1.4 trillion bilingual tokens, accompanied by human preference alignment training. This model successfully achieves several key objectives, including maintaining a smooth conversation flow, imposing minimal deployment requirements, and extending the context length to 32K tokens through the incorporation of FlashAttention.

\textbf{5) BayLing:} BayLing \cite{bayling} is a set of advanced language models that excel in English and Chinese text generation, adeptly follow instructions, and engage in multi-turn conversations. It seamlessly operates on a standard 16GB GPU, facilitating users with translation, writing, and creative suggestions, among other tasks. In our evaluation task, BayLing-7B was selected for evaluation.

\textbf{6) Baichuan-7B:} Baichuan-7B \cite{githubGitHubBaichuanincBaichuan7B}, an open-source language model by Baichuan Intelligent Technology, utilizes the Transformer architecture with 7 billion parameters, trained on about 1.2 trillion tokens. It excels in both Chinese and English, featuring a 4096-token context window. It outperforms similar-sized models on standard Chinese and English benchmarks like C-Eval and MMLU.

\textbf{7) Tigerbot-7B-chat:} Tigerbot-7B-chat \cite{TigerBot23TigerResearch}, derived from TigerBot-7B-base, underwent fine-tuning with 20M data across various tasks for directives (SFT) and alignment using rejection sampling (RS-HIL). Across 13 key assessments in English and Chinese, it outperforms Llama2-Chat-7B by 29\%, showcasing superior performance compared to similar open-source models worldwide.

\textbf{8) Chinese-LLaMA-Alpaca-2:} 
Chinese-LLaMA-Alpaca-2 \cite{Chinese-LLaMA-Alpaca}, an evolution of Meta's Llama2, marks the second iteration of the Chinese LLaMA \& Alpaca LLM initiative. We've made the Chinese LLaMA-2 (the base model) and Alpaca-2 (the instruction-following model) open source. These models have been enhanced and refined with a broader Chinese vocabulary compared to the original Llama2. It conducted extensive pre-training with abundant Chinese data, significantly bolstering our grasp of Chinese language semantics, resulting in a substantial performance boost compared to the first-gen models. The standard version supports a 4K context, while the long-context version accommodates up to 16K context. Moreover, all models can expand their context size using the NTK method, going beyond 24K+.

\textbf{9) Chinese-Falcon-7B:} Chinese-Falcon-7B \cite{githubGitHubCVISZULinly}, developed by the Linly team, expands the Chinese vocabulary of the Falcon model and transfers its language capabilities to Chinese through Chinese and Chinese-English parallel incremental pre-training. Pre-training was conducted using 50GB of data, with 20GB of general Chinese corpora providing Chinese language proficiency and knowledge to the model, 10GB of Chinese-English parallel corpora aligning the model's Chinese and English representations to transfer English language proficiency to Chinese, and 20GB of English corpora used for data replay to mitigate model forgetting.

\textbf{10) iFLYTEK Spark (V2.0):} iFLYTEK Spark (V2.0) \cite{spark} is a next-generation cognitive intelligent model that possesses interdisciplinary knowledge and language comprehension capabilities. It can understand and execute tasks based on natural conversation. It offers functionalities such as multimodal interaction, coding abilities, text generation, mathematical capabilities, language comprehension, knowledge answering, and logical reasoning. Its API enables applications to quickly access interdisciplinary knowledge and powerful natural language understanding capabilities, effectively addressing pressing issues in specific contexts.

For SFT, Llama2-7B, Llama2-7B-Chat, Llama2-Chinese-7B-Chat, and ChatGLM2-6B employed identical training and LoRA configuration settings. For inference, the anterior models with their SFT versions and the listed models above conformed to the same inference configuration. Experiments have been performed to validate the effectiveness of ChatRadio-Valuer on the multi-institution and multi-system radiology report generation.

\subsection{Generalization Performance Evaluation Across Institutions}

Firstly, to assess the ChatRadio-Valuer's generalizable abilities to transfer its acquired knowledge and skills to diverse institutions or sources, we conduct extensive cross-institution performance comparisons involving 15 models. As shown in Table \ref{tab:Cross-institution comparison(Part 1)} and Table \ref{tab:Cross-institution comparison(Part 2)} for details, ChatRadio-Valuer's evaluation encompasses treating data from Institutions 2 through 6 as external sources, revealing a noteworthy degree of generalizability and similarity with the data originating from Institution 1. Our results underscore significant advantages of ChatRadio-Valuer over other models, particularly evident in its absolute superiority in R-1, R-2, and R-L scores on datasets from Institutions 2 and 5. Furthermore, ChatRadio-Valuer secures the highest R-2 score on the dataset from Institution 6. Although it does not attain the top score across all metrics for Institutions 3, 4, and 6, ChatRadio-Valuer consistently maintains a prominent position within the first tier. Taking into account influential external factors like hardware and environmental disparities, we can assert that ChatRadio-Valuer demonstrates exceptional cross-institution performance.

\begin{table}[!t]
\centering
\caption{Cross-institution comparison (Part 1). Results of data from Institution 1 - 3 are shown. For results within each institution, each model corresponds to three similarity scores, R-1, R-2, and R-L.}
\label{tab:Cross-institution comparison(Part 1)}
\resizebox{\textwidth}{!}{%
\begin{tabular}{|c|c|c|c|c|c|c|c|c|c|c|c|c|c|c|c|}
\hline
\multirow{2}{*}{Model} & \multicolumn{3}{c|}{Institution 1} & \multicolumn{3}{c|}{Institution 2} & \multicolumn{3}{c|}{Institution 3} \\
\cline{2-10}
 & R-1 & R-2 & R-L & R-1 & R-2 & R-L & R-1 & R-2 & R-L \\
\hline
Llama2-7B & 0.0909 & 0.0402 & 0.0879 & 0.0825 & 0.0344 & 0.0784 & 0.0172 & 0.0062 & 0.0169 \\
Llama2-7B-ft & 0.4324 & 0.2680 & 0.4181 & 0.4657 & 0.2675 & 0.4489 & \textbf{0.0875} & 0.0250 & \textbf{0.0847} \\
Llama2-Chat-7B & 0.1983 & 0.0810 & 0.1885 & 0.1871 & 0.0478 & 0.1780 & 0.0178 & 0.0040 & 0.0168 \\
ChatRadio-Valuer & \textbf{0.4619} & \textbf{0.2872} & \textbf{0.4464} & \textbf{0.4807} & \textbf{0.2684} & \textbf{0.4608} & 0.0673 & 0.0191 & 0.0644 \\
Llama2-Chinese-Chat-7B & 0.1701 & 0.0751 & 0.1633 & 0.1411 & 0.0386 & 0.1352 & 0.0376 & 0.0127 & 0.0358 \\
Llama2-Chinese-Chat-7B-ft & 0.3221 & 0.1911 & 0.3106 & 0.4045 & 0.2250 & 0.3869 & 0.0675 & 0.0166 & 0.0647 \\
ChatGLM2-ft & 0.2639 & 0.1052 & 0.2556 & 0.2733 & 0.0818 & 0.2578 & 0.0458 & 0.0114 & 0.0441 \\
GPT-3.5-Turbo & 0.1411 & 0.0615 & 0.1341 & 0.0721 & 0.0221 & 0.0679 & 0.0644 & 0.0346 & 0.0623 \\
GPT-4 & 0.1080 & 0.0451 & 0.1029 & 0.0689 & 0.0215 & 0.0646 & 0.0752 & \textbf{0.0393} & 0.0729 \\
BayLing-7B & 0.2732 & 0.1128 & 0.2604 & 0.2643 & 0.0729 & 0.2519 & 0.0412 & 0.0112 & 0.0380 \\
Baichuan-7B & 0.1264 & 0.0578 & 0.1217 & 0.0846 & 0.0247 & 0.0816 & 0.0532 & 0.0234 & 0.0508 \\
Tigerbot-7B-chat-v3 & 0.1338 & 0.0587 & 0.1283 & 0.0783 & 0.0285 & 0.0750 & 0.0817 & 0.0260 & 0.0789 \\
Chinese-Alpaca-2-7B & 0.1884 & 0.0776 & 0.1800 & 0.1837 & 0.0476 & 0.1758 & 0.0546 & 0.0214 & 0.0522 \\
Chinese-Falcon-7B & 0.0770 & 0.0356 & 0.0742 & 0.0360 & 0.0127 & 0.0350 & 0.0391 & 0.0166 & 0.0382 \\
iFLYTEK Spark (V2.0) & 0.1186 & 0.0513 & 0.1142 & 0.0962 & 0.0254 & 0.0918 & 0.0642 & 0.0305 & 0.0622 \\
\hline
\end{tabular}%
}
\end{table}

\begin{table}[!htbp]
\centering
\caption{Cross-institution comparison (Part 2). Results of data from Institution 4 - 6 are shown. For results within each institution, each model corresponds to three similarity scores, R-1, R-2, and R-L.}
\label{tab:Cross-institution comparison(Part 2)}
\resizebox{\textwidth}{!}{%
\begin{tabular}{|c|c|c|c|c|c|c|c|c|c|c|c|c|c|c|c|}
\hline
\multirow{2}{*}{Model} & \multicolumn{3}{c|}{Institution 4} & \multicolumn{3}{c|}{Institution 5} & \multicolumn{3}{c|}{Institution 6} \\
\cline{2-10}
 & R-1 & R-2 & R-L & R-1 & R-2 & R-L & R-1 & R-2 & R-L \\
\hline
Llama2-7B & 0.0028 & 0.0004 & 0.0028 & 0.0441 & 0.0138 & 0.0437 & 0.0122 & 0.0025 & 0.0120 \\
Llama2-7B-ft & \textbf{0.1368} & \textbf{0.0034} & \textbf{0.1368} & 0.1894 & 0.0715 & 0.1836 & 0.0745 & 0.0091 & 0.0715 \\
Llama2-Chat-7B & 0.0066 & 0.0011 & 0.0066 & 0.1184 & 0.0409 & 0.1153 & 0.0369 & 0.0036 & 0.0350 \\
ChatRadio-Valuer & 0.0827 & 0.0000 & 0.0827 & \textbf{0.2195} & \textbf{0.0835} & \textbf{0.2127} & 0.0824 & \textbf{0.0126} & 0.0779 \\
Llama2-Chinese-Chat-7B & 0.0124 & 0.0000 & 0.0124 & 0.1192 & 0.0419 & 0.1150 & 0.0356 & 0.0092 & 0.0345 \\
Llama2-Chinese-Chat-7B-ft & 0.0950 & 0.0000 & 0.0950 & 0.1787 & 0.0642 & 0.1741 & \textbf{0.0915} & 0.0093 & \textbf{0.0896} \\
ChatGLM2-ft & 0.0982 & 0.0007 & 0.0982 & 0.1380 & 0.0304 & 0.1336 & 0.0708 & 0.0044 & 0.0693 \\
GPT-3.5-Turbo & 0.0100 & 0.0000 & 0.0100 & 0.1043 & 0.0664 & 0.0994 & 0.0054 & 0.0000 & 0.0043 \\
GPT-4 & 0.0000 & 0.0000 & 0.0000 & 0.0755 & 0.0313 & 0.0697 & 0.0107 & 0.0000 & 0.0087 \\
BayLing-7B & 0.0108 & 0.0008 & 0.0108 & 0.1649 & 0.0598 & 0.1604 & 0.0677 & 0.0083 & 0.0642 \\
Baichuan-7B & 0.0061 & \textbf{0.0034} & 0.0061 & 0.1083 & 0.0377 & 0.1051 & 0.0329 & 0.0107 & 0.0312 \\
Tigerbot-7B-chat-v3 & 0.0121 & 0.0014 & 0.0121 & 0.1200 & 0.0328 & 0.1159 & 0.0784 & 0.0091 & 0.0782 \\
Chinese-Alpaca-2-7B & 0.0140 & 0.0011 & 0.0140 & 0.1274 & 0.0430 & 0.1230 & 0.0616 & 0.0082 & 0.0600 \\
Chinese-Falcon-7B & 0.0124 & 0.0000 & 0.0124 & 0.0447 & 0.0173 & 0.0432 & 0.0135 & 0.0005 & 0.0135 \\
iFLYTEK Spark (V2.0) & 0.0169 & \textbf{0.0034} & 0.0169 & 0.0756 & 0.0229 & 0.0721 & 0.0355 & 0.0062 & 0.0344 \\
\hline
\end{tabular}%
}
\end{table}

This remarkable capacity for generalization across institutions significantly diminishes the model's reliance on specific institutional data, thereby presenting promising advantages for data sharing and collaboration among institutions. The ability to seamlessly transfer and apply the model across institutions without necessitating retraining positions ChatRadio-Valuer as a preferred tool for numerous medical establishments. The potential value lies in its capability to enhance the quality of patient care, expedite disease diagnosis, and streamline clinical decision-making processes.

Explore further, in order to evaluate the ChatRadio-Valuer's generalizable ability to extend its knowledge and skills to multiple body systems within Institution 1, we conduct performance tests involving 15 models on the designated testing dataset from Institution 1. This testing dataset encompasses five distinct systems (chest, abdomen, skeletal muscle, head, and maxillofacial $\&$ neck parts). As listed in Table \ref{tab:Multiple system similarity (institution 1 only)(Part 1)} and \ref{tab:Multiple system similarity (institution 1 only)(Part 2)} for details, we calculate the similarity for each system among the model-generated diagnosis results (impression) and the professional diagnosis results (REPORT\_RESULT) provided by doctors, yielding R-1, R-2, and R-L values. The outcomes clearly showcase ChatRadio-Valuer's remarkable advantages. Across three systems—abdomen, head, and maxillofacial $\&$ neck—ChatRadio-Valuer achieves the highest scores for R-1, R-2, and R-L values. In the chest system's testing results, both R-1 and R-L values attain the highest scores as well. The remaining test results exhibit remarkable proximity to the maximum values, presenting significantly superior performance compared to the other two fine-tuned models. They firmly belong to the top tier. 

\begin{table}[!ht]
\centering
\caption{Multiple system similarity from Institution 1 (Part 1). The outcomes of data collected from systems 1 through 3 (i.e., Chest to Muscle-skeleton) within Institution 1 are presented herein. Each model is associated with three distinct similarity scores, denoted as R-1, R-2, and R-L.}
\label{tab:Multiple system similarity (institution 1 only)(Part 1)}
\resizebox{\textwidth}{!}{%
\begin{tabular}{|c|c|c|c|c|c|c|c|c|c|c|c|c|c|c|c|}
\hline
\multirow{2}{*}{Model} & \multicolumn{3}{c|}{Chest} & \multicolumn{3}{c|}{Abdomen} & \multicolumn{3}{c|}{Muscle-skeleton} \\
\cline{2-10}
 & R-1 & R-2 & R-L & R-1 & R-2 & R-L & R-1 & R-2 & R-L \\
\hline
Llama2-7B & 0.0993 & 0.0390 & 0.0946 & 0.0934 & 0.0346 & 0.0917 & 0.1307 & 0.0744 & 0.1257 \\
Llama2-7B-ft & 0.5358 & \textbf{0.3369} & 0.5171 & 0.4851 & 0.3033 & 0.4709 & \textbf{0.4763} & \textbf{0.3077} & \textbf{0.4521} \\
Llama2-Chat-7B & 0.2133 & 0.0628 & 0.2024 & 0.2175 & 0.0949 & 0.2092 & 0.2727 & 0.1701 & 0.2507 \\
ChatRadio-Valuer & \textbf{0.5407} & 0.3155 & \textbf{0.5187} & \textbf{0.5283} & \textbf{0.3485} & \textbf{0.5139} & 0.4736 & 0.3035 & 0.4485 \\
Llama2-Chinese-Chat-7B & 0.1524 & 0.0445 & 0.1458 & 0.1662 & 0.0465 & 0.1612 & 0.3568 & 0.2636 & 0.3381 \\
Llama2-Chinese-Chat-7B-ft & 0.4661 & 0.2823 & 0.4476 & 0.3582 & 0.2020 & 0.3484 & 0.4067 & 0.2631 & 0.3885 \\
ChatGLM2-ft & 0.3060 & 0.1071 & 0.2940 & 0.3515 & 0.1472 & 0.3407 & 0.2240 & 0.1304 & 0.2174 \\
GPT-3.5-Turbo & 0.1082 & 0.0270 & 0.1041 & 0.1889 & 0.0659 & 0.1830 & 0.2620 & 0.1822 & 0.2378 \\
GPT-4 & 0.0514 & 0.0055 & 0.0507 & 0.1178 & 0.0220 & 0.1153 & 0.2747 & 0.2012 & 0.2525 \\
BayLing-7B & 0.2652 & 0.0829 & 0.2540 & 0.3414 & 0.1464 & 0.3256 & 0.3520 & 0.2065 & 0.3239 \\
Baichuan-7B & 0.0637 & 0.0172 & 0.0616 & 0.1391 & 0.0404 & 0.1328 & 0.2901 & 0.2038 & 0.2776 \\
Tigerbot-7B-chat-v3 & 0.0934 & 0.0276 & 0.0899 & 0.1394 & 0.0307 & 0.1341 & 0.3388 & 0.2300 & 0.3216 \\
Chinese-Alpaca-2-7B & 0.1645 & 0.0470 & 0.1582 & 0.2308 & 0.0770 & 0.2203 & 0.3396 & 0.2241 & 0.3198 \\
Chinese-Falcon-7B & 0.0383 & 0.0065 & 0.0376 & 0.0758 & 0.0120 & 0.0741 & 0.2420 & 0.1722 & 0.2289 \\
iFLYTEK Spark (V2.0) & 0.0514 & 0.0088 & 0.0503 & 0.1695 & 0.0410 & 0.1629 & 0.3136 & 0.2254 & 0.3001 \\
\hline
\end{tabular}%
}
\end{table}

\begin{table}[!htbp]
\centering
\caption{Multiple system similarity from Institution 1 (Part 2). Outcomes of data collected from systems 4 and 5 (i.e., "Head" and "Maxillofacial \& neck") within Institution 1 are presented herein. Each model is associated with three distinct similarity scores, denoted as R-1, R-2, and R-L.}
\label{tab:Multiple system similarity (institution 1 only)(Part 2)}
\small
\begin{tabular}{|c|c|c|c|c|c|c|}
\hline
\multirow{2}{*}{Model} & \multicolumn{3}{c|}{Head} & \multicolumn{3}{c|}{Maxillofacial $\&$ neck} \\
\cline{2-7}
 & R-1 & R-2 & R-L & R-1 & R-2 & R-L \\
\hline
Llama2-7B & 0.0582 & 0.0293 & 0.0569 & 0.0516 & 0.0234 & 0.0510 \\
Llama2-7B-ft & 0.2276 & 0.1333 & 0.2246 & 0.2548 & 0.1222 & 0.2495 \\
Llama2-Chat-7B & 0.1236 & 0.0406 & 0.1210 & 0.1179 & 0.0478 & 0.1146 \\
ChatRadio-Valuer & \textbf{0.2853} & \textbf{0.1906} & \textbf{0.2825} & \textbf{0.3447} & \textbf{0.1902} & \textbf{0.3396} \\
Llama2-Chinese-Chat-7B & 0.0914 & 0.0325 & 0.0894 & 0.0947 & 0.0364 & 0.0920 \\
Llama2-Chinese-Chat-7B-ft & 0.0868 & 0.0442 & 0.0843 & 0.0762 & 0.0303 & 0.0742 \\
ChatGLM2-ft & 0.1342 & 0.0344 & 0.1319 & 0.2261 & 0.1045 & 0.2226 \\
GPT-3.5-Turbo & 0.0731 & 0.0287 & 0.0699 & 0.0750 & 0.0463 & 0.0750 \\
GPT-4 & 0.0737 & 0.0246 & 0.0695 & 0.0858 & 0.0346 & 0.0858 \\
BayLing-7B & 0.1628 & 0.0516 & 0.1593 & 0.2316 & 0.1187 & 0.2268 \\
Baichuan-7B & 0.0928 & 0.0353 & 0.0906 & 0.1304 & 0.0624 & 0.1279 \\
Tigerbot-7B-chat-v3 & 0.0650 & 0.0287 & 0.0632 & 0.0674 & 0.0217 & 0.0661 \\
Chinese-Alpaca-2-7B & 0.0959 & 0.0327 & 0.0933 & 0.1033 & 0.0420 & 0.1001 \\
Chinese-Falcon-7B & 0.0312 & 0.0154 & 0.0304 & 0.0449 & 0.0200 & 0.0440 \\
iFLYTEK Spark (V2.0) & 0.0514 & 0.0163 & 0.0496 & 0.0466 & 0.0176 & 0.0455 \\
\hline
\end{tabular}
\end{table}

On the other hand, to evaluate on ChatRadio-Valuer’s generalizable ability to extend its knowledge and skills to multiple body systems within external individual institutions, we conduct performance tests on chest, abdomen, skeletal muscle, head, and maxillofacial $\&$ neck parts respectively from Institution 2 to Institution 6. As shown in the columns from the corresponding institutions between Table \ref{tab:Cross-institution comparison(Part 1)} and Table \ref{tab:Cross-institution comparison(Part 2)}, it can be found that in Institution 2 and Institution 5, ChatRadio-Valuer leads in all three metrics: R-1, R-2, and R-L, and ChatRadio-Valuer outperforms in R-2 far from other models in Institution 6 while only slightly inferior to others' in R-1 and R-L, which is similar in Institution 3 and Institution 4. These concrete results are driven by ChatRadio-Valuer's superior generalizability derived from data in Institution 1. ChatRadio-Valuer successfully expand the knowledge acquired from Institution 1 to other idiosyncratic institutions. And as these institutions contain their corresponding disease system purely, in other words, ChatRadio-Valuer predominates in either broadening knowledge from single to multiple unseen datasets or applying this knowledge in addressing multi-system clinical problems.

In summary, these successes unequivocally affirms the following conclusion:
\begin{itemize}
    \item ChatRadio-Valuer showcases exceptional adaptability in handling the intricacies and nuances of radiology reports, enabling it to deliver precise and continuous diagnostic outcomes with superior performance.
    \item ChatRadio-Valuer demonstrates a high degree of generalizability in the diagnosis of multiple body systems, not limited to specific systems but capable of applying to diverse medical data and producing high-quality diagnostic results.
    \item ChatRadio-Valuer exhibits robust generalization capabilities, efficiently handling a wide range of data from many institutions and adapting seamlessly to various medical scenarios.
\end{itemize}

\subsection{Generalization Performance for Multi-system Diseases}

\begin{table}[!b]
\centering
\caption{Five-system comparison across institutions (Part 1). Results derived from the data acquired from systems 1 to 3, specifically labeled as "Chest" to "Muscle-skeleton," are herein disclosed. Each of these systems encompasses data stemming from two different institutions (comprising a blend), and for each model, three unique similarity scores are employed, namely, R-1, R-2, and R-L.}
\label{tab:Five-system comparison across institutions(Part 1)}
\resizebox{\textwidth}{!}{%
\begin{tabular}{|c|c|c|c|c|c|c|c|c|c|c|c|c|c|c|c|}
\hline
\multirow{2}{*}{Model} & \multicolumn{3}{c|}{Chest} & \multicolumn{3}{c|}{Abdomen} & \multicolumn{3}{c|}{Muscle-skeleton} \\
\cline{2-10}
 & R-1 & R-2 & R-L & R-1 & R-2 & R-L & R-1 & R-2 & R-L \\
\hline
Llama2-7B & 0.0921 & 0.0370 & 0.0877 & 0.0885 & 0.0328 & 0.0868 & 0.1258 & 0.0716 & 0.1209 \\
Llama2-7B-ft & 0.5058 & \textbf{0.3072} & 0.4880 & 0.4593 & 0.2892 & 0.4459 & \textbf{0.4633} & 0.2960 & \textbf{0.4399} \\
Llama2-Chat-7B & 0.2021 & 0.0564 & 0.1920 & 0.2045 & 0.0890 & 0.1967 & 0.2625 & 0.1636 & 0.2414 \\
ChatRadio-Valuer & \textbf{0.5151} & 0.2954 & \textbf{0.4939} & \textbf{0.4983} & \textbf{0.3271} & \textbf{0.4847} & 0.4586 & \textbf{0.2981} & 0.4345 \\
Llama2-Chinese-Chat-7B & 0.1476 & 0.0420 & 0.1413 & 0.1578 & 0.0443 & 0.1531 & 0.3436 & 0.2535 & 0.3256 \\
Llama2-Chinese-Chat-7B-ft & 0.4398 & 0.2578 & 0.4216 & 0.3393 & 0.1900 & 0.3300 & 0.3948 & 0.2531 & 0.3773 \\
ChatGLM2-ft & 0.2920 & 0.0963 & 0.2785 & 0.3317 & 0.1383 & 0.3214 & 0.2192 & 0.1254 & 0.2128 \\
GPT-3.5-Turbo & 0.0812 & 0.0233 & 0.0770 & 0.1192 & 0.0484 & 0.1154 & 0.1188 & 0.0787 & 0.1084 \\
GPT-4 & 0.0645 & 0.0174 & 0.0611 & 0.0940 & 0.0317 & 0.0916 & 0.0924 & 0.0676 & 0.0849 \\
BayLing-7B & 0.2648 & 0.0786 & 0.2531 & 0.3219 & 0.1376 & 0.3069 & 0.3389 & 0.1986 & 0.3119 \\
Baichuan-7B & 0.0726 & 0.0204 & 0.0702 & 0.1335 & 0.0393 & 0.1275 & 0.2792 & 0.1962 & 0.2672 \\
Tigerbot-7B-chat-v3 & 0.0870 & 0.0280 & 0.0835 & 0.1357 & 0.0304 & 0.1306 & 0.3262 & 0.2212 & 0.3098 \\
Chinese-Alpaca-2-7B & 0.1727 & 0.0472 & 0.1657 & 0.2194 & 0.0733 & 0.2094 & 0.3271 & 0.2155 & 0.3080 \\
Chinese-Falcon-7B & 0.0373 & 0.0091 & 0.0365 & 0.0734 & 0.0123 & 0.0718 & 0.2332 & 0.1656 & 0.2206 \\
iFLYTEK Spark (V2.0) & 0.0706 & 0.0159 & 0.0681 & 0.1626 & 0.0403 & 0.1564 & 0.3022 & 0.2169 & 0.2892 \\
\hline
\end{tabular}%
}
\end{table}

\begin{table}[!htbp]
\centering
\caption{Five-system comparison across institutions (Part 2). Results derived from the data acquired from systems 4 and 5, specifically labeled as "Head" to "Maxillofacial \& neck," are herein disclosed. Each of these systems encompasses data stemming from two different institutions (comprising a blend), and for each model, three unique similarity scores are employed, namely, R-1, R-2, and R-L.}
\label{tab:Five-system comparison across institutions(Part 2)}
\small
\begin{tabular}{|c|c|c|c|c|c|c|}
\hline
\multirow{2}{*}{Model} & \multicolumn{3}{c|}{Head} & \multicolumn{3}{c|}{Maxillofacial $\&$ neck} \\
\cline{2-7}
 & R-1 & R-2 & R-L & R-1 & R-2 & R-L \\
\hline
Llama2-7B & 0.0573 & 0.0284 & 0.0561 & 0.0477 & 0.0213 & 0.0472 \\
Llama2-7B-ft & 0.2252 & 0.1295 & 0.2221 & 0.2370 & 0.1110 & 0.2319 \\
Llama2-Chat-7B & 0.1233 & 0.0406 & 0.1206 & 0.1099 & 0.0434 & 0.1067 \\
ChatRadio-Valuer & \textbf{0.2812} & \textbf{0.1840} & \textbf{0.2782} & \textbf{0.3188} & \textbf{0.1727} & \textbf{0.3138} \\
Llama2-Chinese-Chat-7B & 0.0931 & 0.0331 & 0.0910 & 0.0888 & 0.0337 & 0.0863 \\
Llama2-Chinese-Chat-7B-ft & 0.0925 & 0.0454 & 0.0899 & 0.0777 & 0.0282 & 0.0757 \\
ChatGLM2-ft & 0.1344 & 0.0341 & 0.1320 & 0.2108 & 0.0946 & 0.2075 \\
GPT-3.5-Turbo & 0.0917 & 0.0512 & 0.0875 & 0.0238 & 0.0123 & 0.0231 \\
GPT-4 & 0.0748 & 0.0286 & 0.0696 & 0.0305 & 0.0092 & 0.0291 \\
BayLing-7B & 0.1629 & 0.0521 & 0.1594 & 0.2154 & 0.1078 & 0.2108 \\
Baichuan-7B & 0.0938 & 0.0355 & 0.0915 & 0.1207 & 0.0573 & 0.1183 \\
Tigerbot-7B-chat-v3 & 0.0684 & 0.0289 & 0.0664 & 0.0684 & 0.0204 & 0.0673 \\
Chinese-Alpaca-2-7B & 0.0978 & 0.0333 & 0.0952 & 0.0992 & 0.0386 & 0.0961 \\
Chinese-Falcon-7B & 0.0320 & 0.0156 & 0.0312 & 0.0418 & 0.0181 & 0.0410 \\
iFLYTEK Spark (V2.0) & 0.0529 & 0.0167 & 0.0509 & 0.0455 & 0.0165 & 0.0444 \\
\hline
\end{tabular}
\end{table}

To further explore the effectiveness and generalization of the ChatRadio-Valuer to broaden its knowledge and skills to multiple body systems from the whole institutions, we conduct experiments on data from all the five distinct systems (i.e. chest, abdomen, muscle-skeleton, head, maxillofacial $\&$ neck), with the experimental results reported in Table \ref{tab:Five-system comparison across institutions(Part 1)}, and Table \ref{tab:Five-system comparison across institutions(Part 2)}. This assessment involves 15 models and aims to gauge the model's adaptability and effectiveness in handling diverse disease-related information. We can obtain several observations from the results. First, ChatRadio-Valuer outperforms its baselines on  R-1, R-2, and R-L metrics across selected LLMs, regardless of the system considered (i.e., chest, abdomen, muscle-skeleton, head, maxillofacial $\&$ neck). The significant performance demonstrates the model's generalization, particularly on disease data from the abdomen, head, and maxillofacial $\&$ neck. While ChatRadio-Valuer does not show the best performance across all metrics for Chest and muscle-skeleton data, it occupies a prominent position within the first tier of performance (refer to Table \ref{tab:Five-system comparison across institutions(Part 1)} for detailed metrics). This consistent performance highlights the model's reliability in processing multi-disease data. Second, taking into account influential external factors like hardware and environmental disparities, we can confidently assert that ChatRadio-Valuer demonstrates exceptional cross-system performance. This great generalization performance across diseases significantly reduces the model's reliance on disease-specific data, providing insights and suggestions for rare diseases and comprehensive diagnosis of disease. The potential value of ChatRadio-Valuer is to improve comprehensive diagnostic quality, and streamline clinical decision-making processes. Finally, combining the results of Table \ref{tab:Multiple system similarity (institution 1 only)(Part 1)}, \ref{tab:Multiple system similarity (institution 1 only)(Part 2)}, \ref{tab:Five-system comparison across institutions(Part 1)}, and \ref{tab:Five-system comparison across institutions(Part 2)}, it can be found that the ChatRadio-Valuer can not only distinguish various diseases in a certain institution, but also distinguish multiple diseases in mixed institutions, which demonstrates the adaptability in learning distinct features of diseases information. Distinguishing and generating related information across different disease contexts, shows its significant promise in multi-system data analysis.

\begin{itemize}
    \item ChatRadio-Valuer is proficient at simultaneously managing data from diverse medical institutions and disease systems, thereby offering a spectrum of complementary insights from various perspectives. This capability enriches our comprehensive comprehension of diseases and expedites the generation of results that provide healthcare professionals with profound analyses and insights.
    \item ChatRadio-Valuer demonstrates adeptness in acquiring distinct diagnostic outcomes and diagnostic styles from multi-institutional and multi-system datasets. Throughout the fine-tuning process, it perpetually fine-tunes and hones its knowledge, culminating in more precise and succinct articulations. This holds considerable clinical promise for the practical utilization of the generated results.
    \item ChatRadio-Valuer synergistically leverages the strengths of both dimensions, rendering it applicable in a broader array of contexts. It possesses the capacity to process data from diverse healthcare institutions and across distinct organ systems, thereby augmenting its versatility for broader applications in the realms of medicine and biomedical engineering.
\end{itemize}

\subsection{Expert Evaluation}\label{sec:Expert Evaluation}
The feasibility on ChatRadio-Valuer for radiology report generation has been fully verified in Table \ref{tab:Cross-institution comparison(Part 1)} - Table \ref{tab:Five-system comparison across institutions(Part 2)}, which has proven its availability in smoothly presenting semantic information. However, further exploration of these results to determine their clinical value is necessarily required. Therefore, taking into account the practical clinical needs from both a holistic and specific data distribution perspective, we conduct a comprehensive analysis and study on the diagnostic effectiveness of different system diseases in different institutions. And the practical clinical evaluation metrics proposed by radiologists are shown in Figure \ref{fig5}.  

\begin{figure}[!h]
    \centering
    \includegraphics[width=1\textwidth]{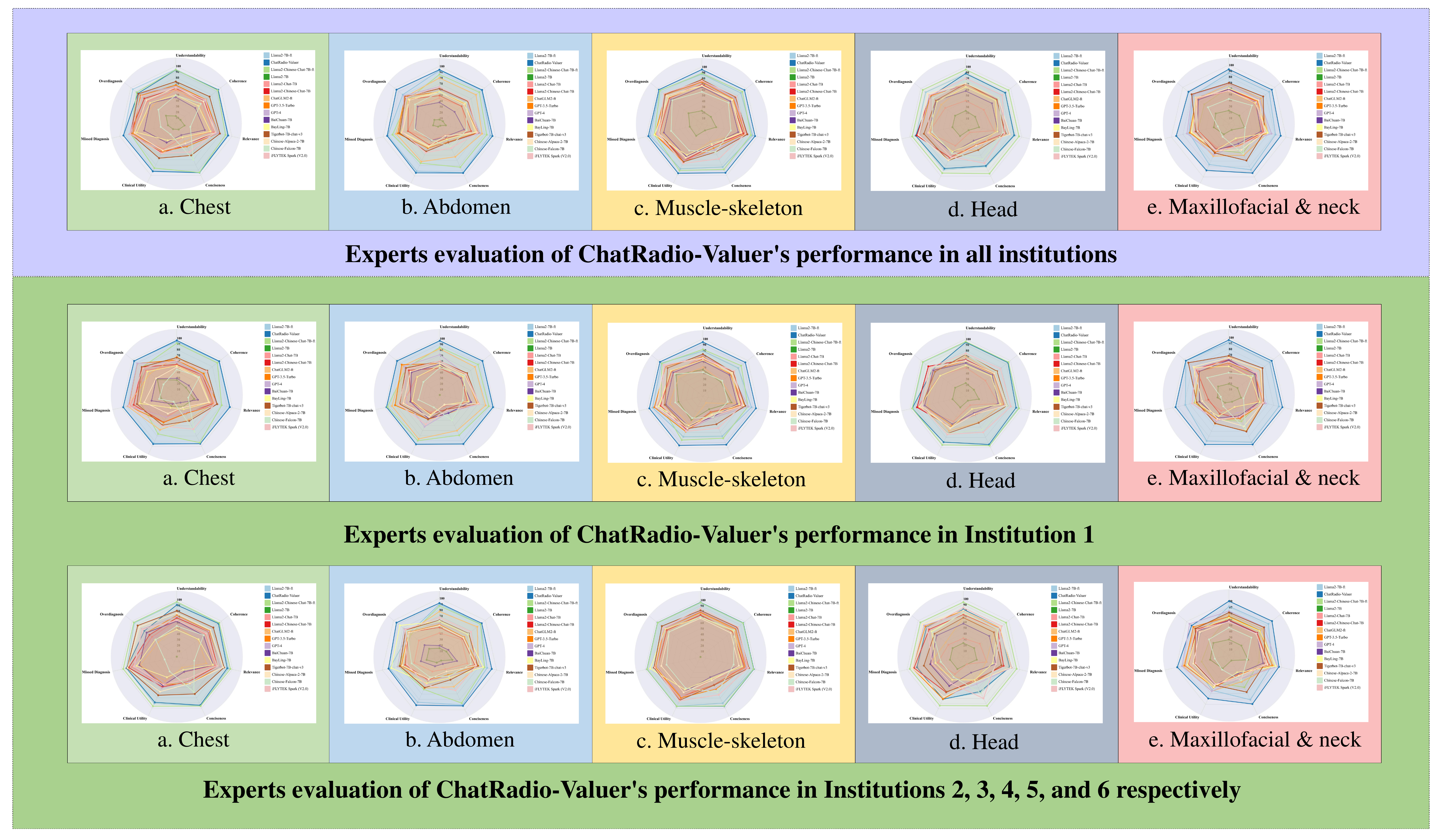}
    \caption{Expert evaluation results on clinical metrics. The first row represents the overall clinical generalizability (OG) across various systems with the whole institutions data. Results between the second row and the third row appraise models' in-house generalizability (IHG), and out-of-house generalizability (OHG), respectively.}
    \label{fig:expertEval}
\end{figure}

The evaluation results, including junior, intermediate, and senior radiologists, are performed to take the average operation, which are shown in Figure \ref{fig:expertEval}. All evaluations are based on the five systems: chest, abdomen, muscle-skeleton, head, and maxillofacial $\&$ neck. In each plot (e.g., "a. Chest", "b. Abdomen", "c. Musculoskeletal", etc.), the seven clinical metrics are uniformly distributed in the circle. The first row represents the evaluation on the scale of the whole testing dataset, including the dataset in the second row and third row. This result comprehensively evaluates the overall generalizability (OG) across the whole institutions and systems. The second row refers to the evaluation on the specific dataset of Institution 1, which directly indicates the performance of our strategy of fine-tuning and the generalizability within Institution 1, namely the in-house generalizability (IHG). The final row, represents the evaluation of the testing dataset across institutions 2, 3, 4, 5 and 6, respectively. Since institutions 2, 3, 4, 5 and 6 are irrelevant to the training phase, the performance in this row more comprehensively evaluates our method's adaptability and generalizability across data from different institutions with different writing styles, defined by out-of-house generalizability (OHG). To evaluate in detail, the expanding area of each plot indicates the clinical performance of models. In other words, the larger the covered area, the better the performance.

Generally speaking, it can be demonstrated that ChatRadio-Valuer covers the most amounts of area in almost all subplots, across the three scales of the dataset. Within each row, ChatRadio-Valuer's leading performance in the five systems—chest, abdomen, muscle-skeleton, head, and maxillofacial $\&$ neck-shows its versatility across different systems. In other words, for example, the OG, IHG, and OHG on the chest system (the first column "a. Chest" of the three rows) remain the same superior level. Expanding to the five systems, in the three rows, OG, IHG, and OHG respectively, ChatRadio-Valuer obtains excellent scores though the scores in "Head" are only slightly inferior to those in other systems (probably due to the written styles are hard to recognize for LLM). This reveals ChatRadio-Valuer's great OHG. Notice that the best performance in "Head" is occupied by Llama2-Chinese-Chat-7B-ft which adopts our training strategy. These phenomena collaboratively prove the success of the fine-tuning strategy and the advance of our proposed ChatRadio-Valuer. Its superior generalizability sheds light on further real clinical radiology applications over even such complex and heterogeneous cases.


\subsection{Practical Utility Assessment}

In this section, we employ assessment on ChatRadio-Valuer's practical utility shown in Table \ref{tab:model-comparison}, involving three metrics: parameter count, fine-tuning time, and testing time. In the condition of the same amount of parameters, fine-tuning time evaluates how efficient for LLMs to achieve the desired level of model performance. Similarly, testing time indicates how long it takes to make a causal inference. Note that for doctors at the bottom of Table \ref{tab:model-comparison}, the testing time refers to the time for them to derive a comprehensive impression from the corresponding report.


\begin{table}[!htbp]
\centering
\caption{Practical utility evaluation on the proposed ChatRadio-Valuer compared with SOTA LLMs.}
\label{tab:model-comparison}
\resizebox{\textwidth}{!}{%
\begin{tabular}{|l|c|c|c|}
\hline
\rowcolor{lightgray}
\textbf{Model} & \textbf{Parameter Count} & \textbf{Fine-tuning Time (h)} & \textbf{Testing Time (s)} \\
\hline
Llama2-7B                   & 7B         & 54                & 17.9893 \\
Llama2-7B-ft               & 7B         & 54                & 22.5722 \\
Llama2-Chat-7B             & 7B         & 54                & 8.9997  \\
ChatRadio-Valuer         & 7B         & 54                & 22.3577 \\
Llama2-Chinese-Chat-7B    & 7B         & 54                 & 7.8787  \\
Llama2-Chinese-Chat-7B-ft & 7B         & 54                 & 16.2844 \\
ChatGLM2-ft                 & 6B         & 28                 & 7.8340  \\
GPT-3.5-Turbo               & 175B    & NA                 & 1.2081  \\
GPT-4                        & 1.8T    & NA                 & 2.0228  \\
BayLing-7B                  & 7B         & NA                 & 12.8729 \\
Baichuan-7B                 & 7B         & NA                 & 11.4543 \\
Chinese-Falcon-7B           & 7B         & NA                 & 5.8440  \\
Chinese-Alpaca-2-7B         & 7B         & NA                 & 4.9946  \\
Tigerbot-7B-chat-v3             & 7B         & NA                 & 2.7295  \\
iFLYTEK Spark (V2.0)         & $\geq 170\mathrm{B}$    & NA                 & 3.8263  \\
\hline
Doctors         & NA    & NA                 & 60-180  \\
\hline
\end{tabular}%
}
\end{table}


In the situation that most of the models, except the four, are 7B-version, the evaluation on them are only slightly affected by the model size, especially with the same fine-tuning and inferring configuration. We conduct fine-tuning on 4 models (i.e., Llama2-7B, Llama2-7B-Chat, Llama2-Chinese-Chat-7B, and ChatGLM2). The Llama2 family consumes approximately 54 hours for each while the ChatGLM2 only costs 28 hours. For the testing time, Llama2-7B-ft takes the longest time, 22.5722s, and our ChatRadio-Valuer uses 22.3577s for a single causal inference. The fastest models belong to the GPT family, 1.2081s for GPT-3.5-Turbo and 2.0228s for GPT-4 per inference. For doctors, from junior to senior, the diagnosing time ranges from 60s to 180s per sample inference. Supported by Table \ref{tab:Cross-institution comparison(Part 1)} - \ref{tab:Five-system comparison across institutions(Part 2)}, the effectiveness of ChatRadio-Valuer is verified through high similarity scores among other prevailing methods. In comparison to different levels radiologists, ChatRadio-Valuer shows a significant advantage in terms of time efficiency indicated in Table \ref{tab:model-comparison}. These results showcase ChatRadio-Valuer's superior capability in both effectiveness and efficiency.

\subsection{Result Analysis}

\begin{figure}[htbp!]
    \centering
    \includegraphics[width=0.93\textwidth,height=!,keepaspectratio]{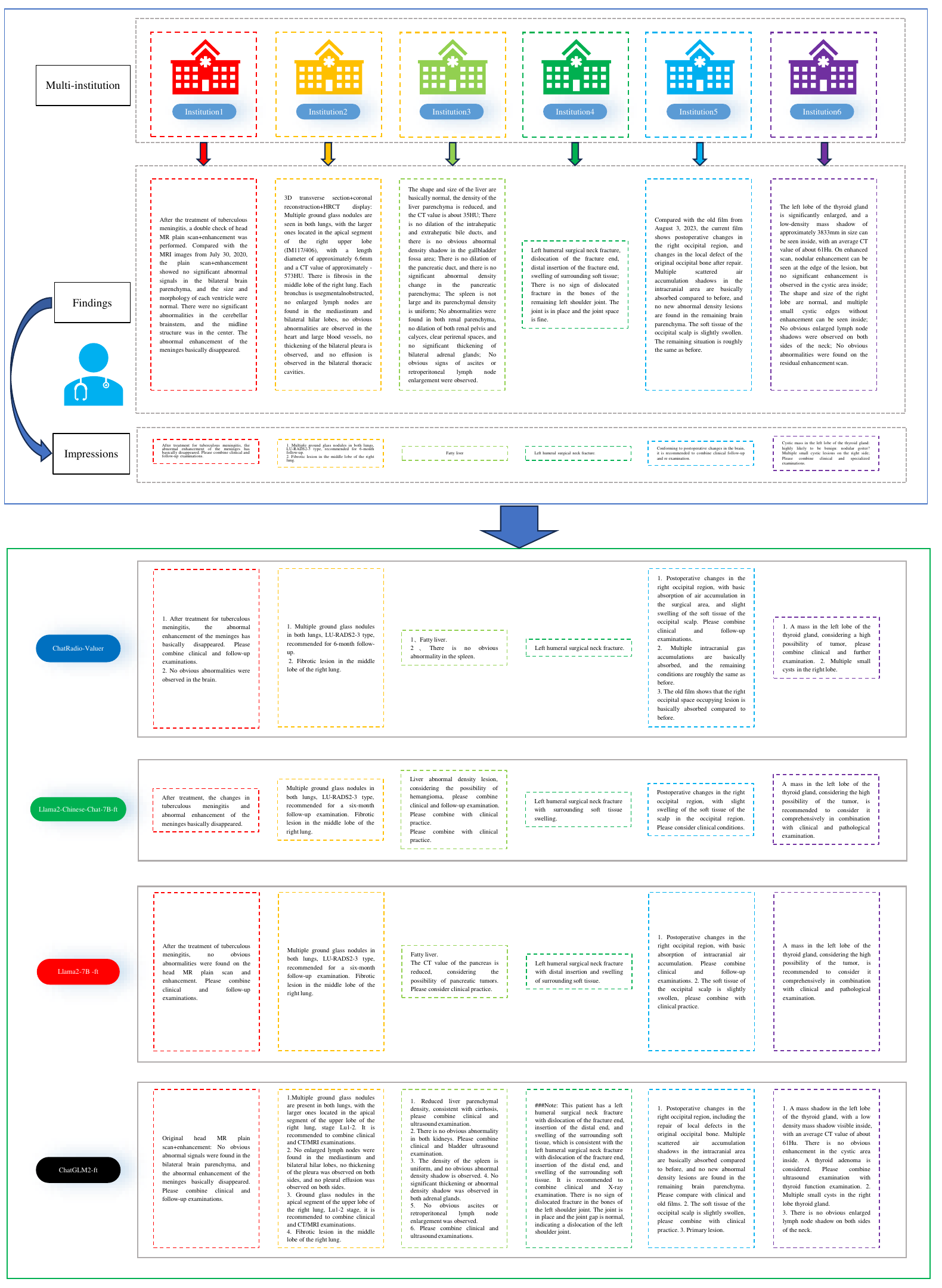}
    \caption{Diagram of the response results among fine-tuned models and state-of-the-art LLMs across institutions (Part 1). Within each institution (containing the institutions from 1 to 6), every model generates corresponding clinical inference based on the same finding.}
    \label{fig:3.1-1}
\end{figure}

\begin{figure}[htbp!]
    \centering
    \includegraphics[width=1.0\textwidth,height=!,keepaspectratio]{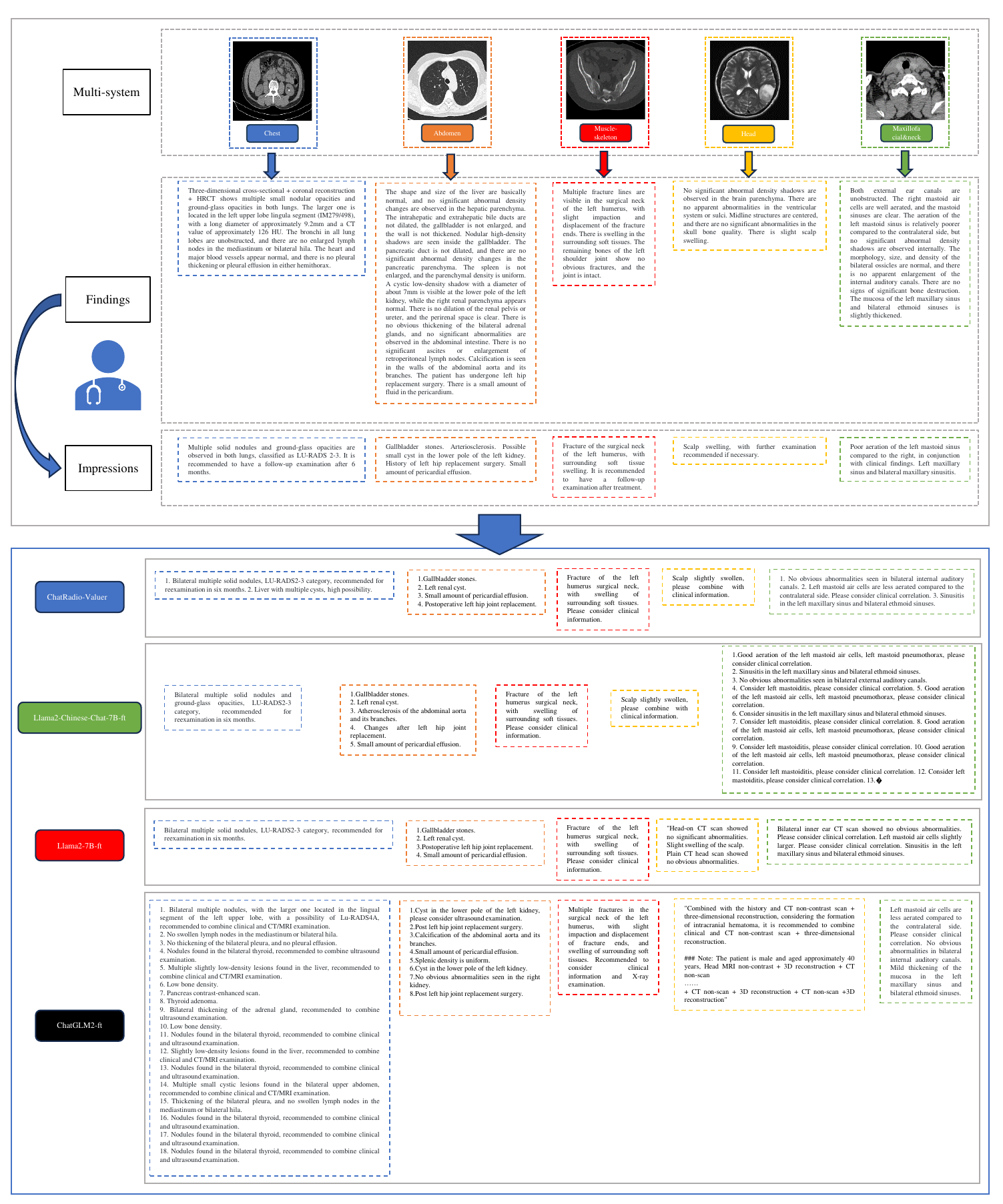}
    \caption{Diagram of the response results among fine-tuned models and state-of-the-art LLMs across systems (Part 1). Within each system (i.e., chest, abdomen, muscle-skeleton, head, and maxillofacial \& neck), each model generates corresponding clinical inference based on the same finding.}
    \label{fig:3.2-1}
\end{figure}

To assess the real-world performance of ChatRadio-Valuer and the SOTA LLMs (including three fine-tuning pairs, one fine-tuned ChatGLM2-6B, prestigious GPT family, etc. shown in Figure \ref{fig:modelpool}) in the domain of radiology report generation respectively across multiple institutions and diverse disease systems, we employ a random sampling approach to select representative data from each institution and across various disease systems for visualization in Figure \ref{fig:3.1-1} - \ref{fig:3.2-1}, and \ref{fig:3.1-2} - \ref{fig:3.2-3}. The illustration of the superiority of ChatRadio-Valuer and our fine-tuning strategy is shown in Figure \ref{fig:3.1-1} and Figure \ref{fig:3.2-1}, respectively. Due the limitation of page size, we attach the rest four diagrams, Figure \ref{fig:3.1-2} - \ref{fig:3.2-3}, in Appendix. Figure \ref{fig:3.1-1}, \ref{fig:3.1-2}, and \ref{fig:3.1-3} are results across institutions. The above blue box illustrates the process that findings from the six institutions are proceeded by doctors to generate the impressions. The green box below is the corresponding inference set constituting outputs from the selected models. Similarly, Figure \ref{fig:3.2-1}, \ref{fig:3.2-2}, and \ref{fig:3.2-3} represent the same but results across systems (i.e., findings are from the five systems rather than institutions). 

For multi-institution comparison, referring to Figure \ref{fig:3.1-1}, \ref{fig:3.1-2}, and \ref{fig:3.1-3} (only take Figure \ref{fig:3.1-1} for example due to simplicity consideration), methods utilizing our fine-tuned strategy (i.e., with the suffix "-ft") have better performance in efficacy and meaningfulness. In terms of the finding and impression pair from Institution 1, the generated textual contents of our fine-tuned models are characterized by increased conciseness and precision, as they effectively reduce extraneous outputs while retaining essential clinical information. Upon a thorough examination of these findings, it is evident that ChatRadio-Valuer produces outcomes distinguished by a more intricate and refined linguistic granularity when compared to other fine-tuned models. Expanding on the aforementioned improvements, it demonstrates an elevated clinical utility. Similarly for multi-system comparison, indicated by Figure \ref{fig:3.2-1}, \ref{fig:3.2-2}, and \ref{fig:3.2-3}, fine-tuned models perform better than other base models. Take the abdomen system in Figure \ref{fig:3.2-1} for example, fine-tuned models generate more succinct and relevant inferences. More significantly, again, ChatRadio-Valuer performs the best among other fine-tuned methods.

\begin{figure*}[htbp!]
    \centering  

    \includegraphics[width=0.78\textwidth,height=1.27\textwidth]{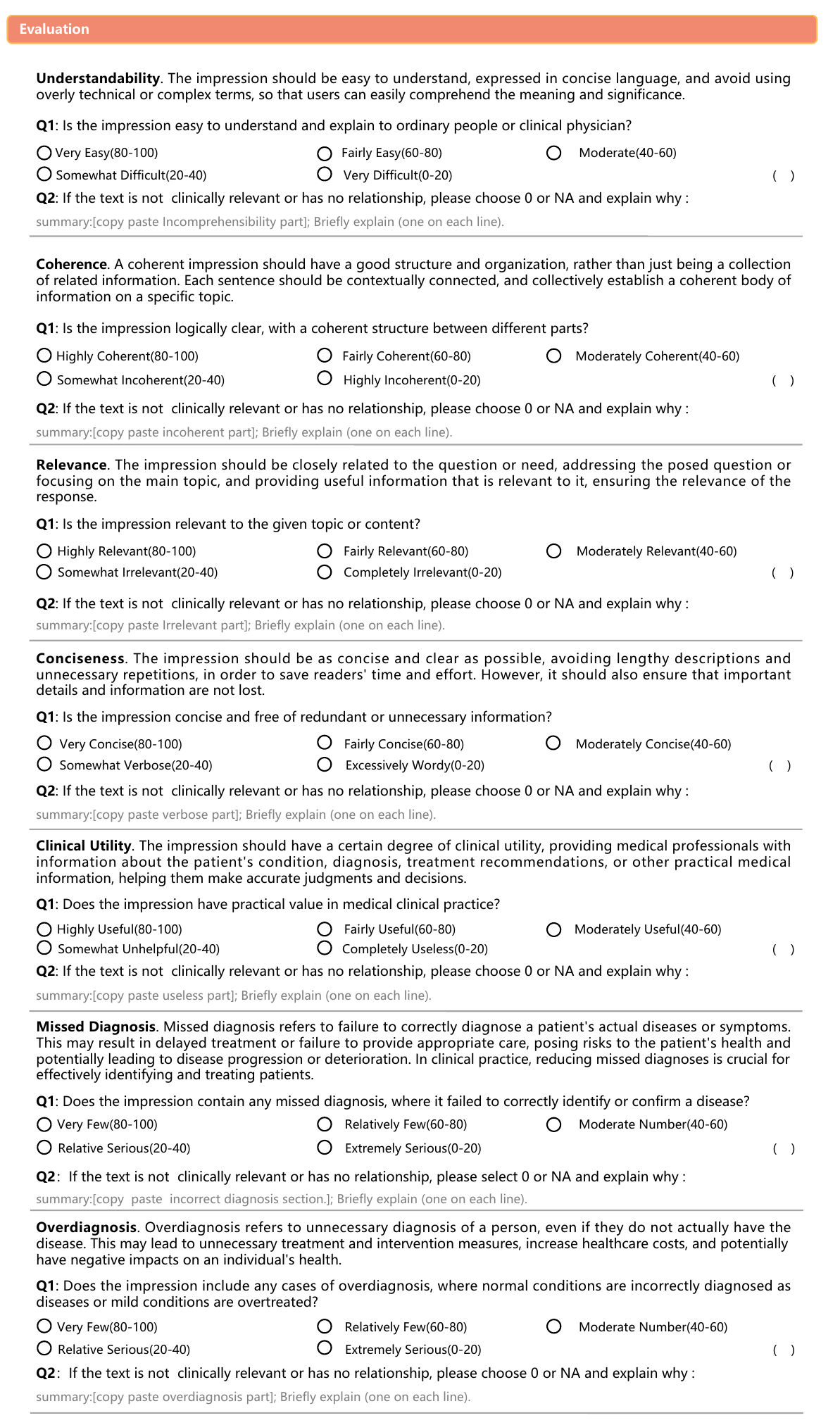}
    \caption{Questionnaire for experts evaluation. Seven metrics (i.e., understandability, coherence, relevance, conciseness, clinical utility, missed Diagnosis, overdiagnosis), are provided for radiologists to comprehensively evaluate the results in clinical usage.}
    \label{fig5}
\end{figure*}

\begin{figure}[htbp!]
    \centering
    \includegraphics[width=0.82\textwidth,height=1.27\textwidth,keepaspectratio]{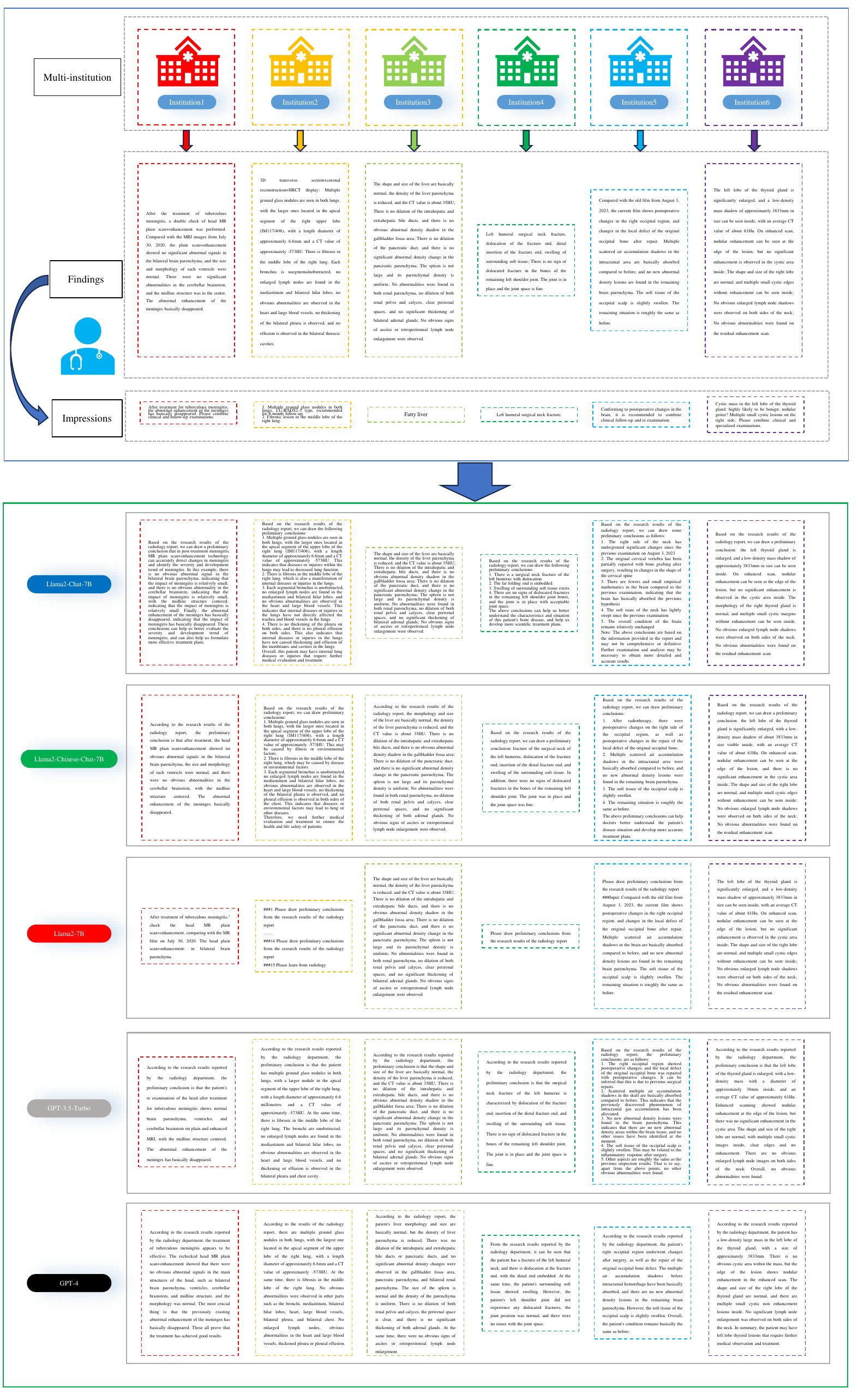}
    \caption{Diagram of the response results among fine-tuned models and state-of-the-art LLMs across institutions (Part 2). Within each institution (containing the institutions from 1 to 6), every model generates corresponding clinical inference based on the same finding.}
    \label{fig:3.1-2}
\end{figure}

\begin{figure}[htbp!]
    \centering
    \includegraphics[width=0.83\textwidth,height=1.27\textwidth]{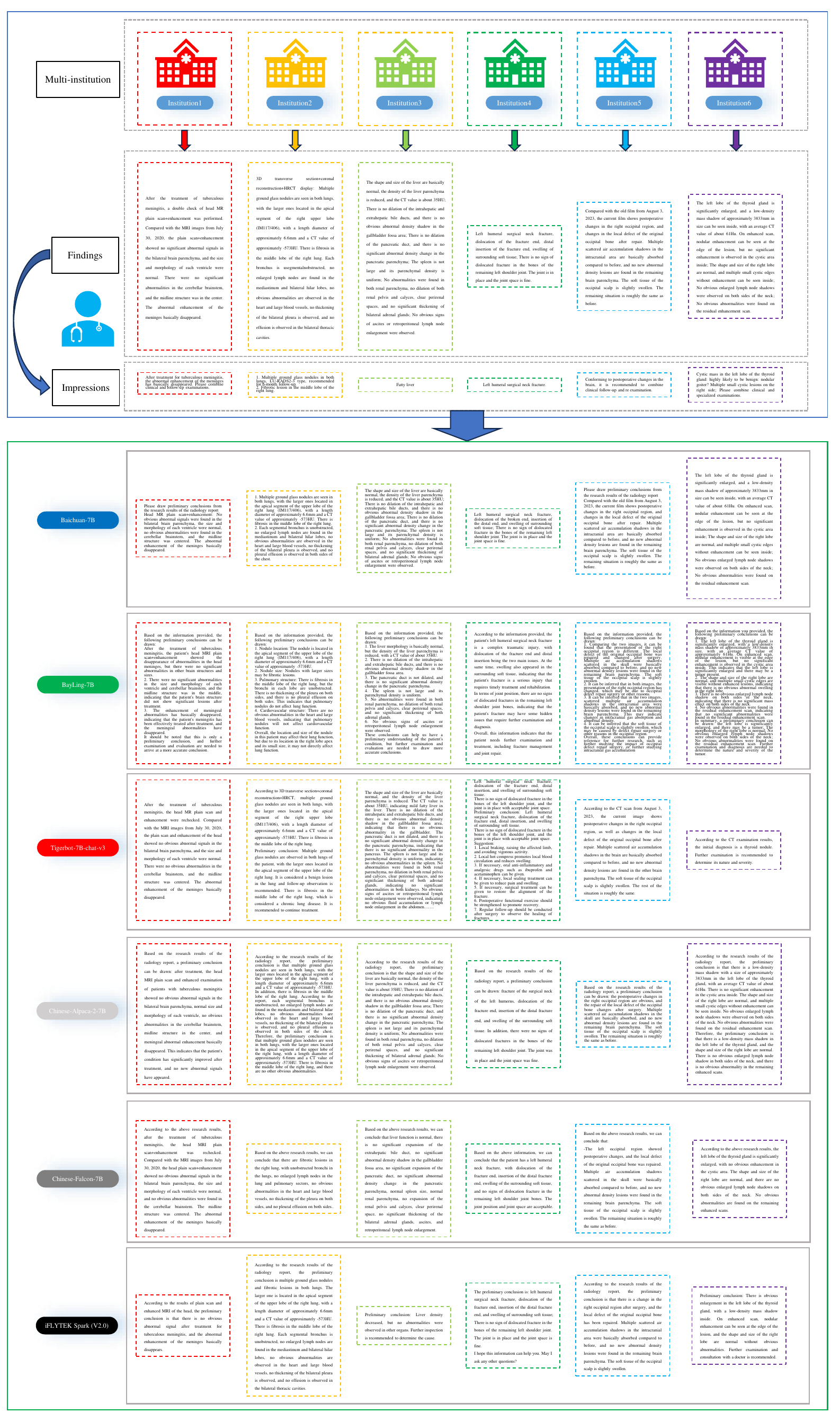}
    \caption{Diagram of the response results among fine-tuned models and state-of-the-art LLMs across institutions (Part 3). Within each institution (containing the institutions from 1 to 6), every model generates corresponding clinical inference based on the same finding.}
    \label{fig:3.1-3}
\end{figure}

\begin{figure}[htbp!]
    \centering

    \includegraphics[width=0.85\textwidth,height=1.24\textwidth]{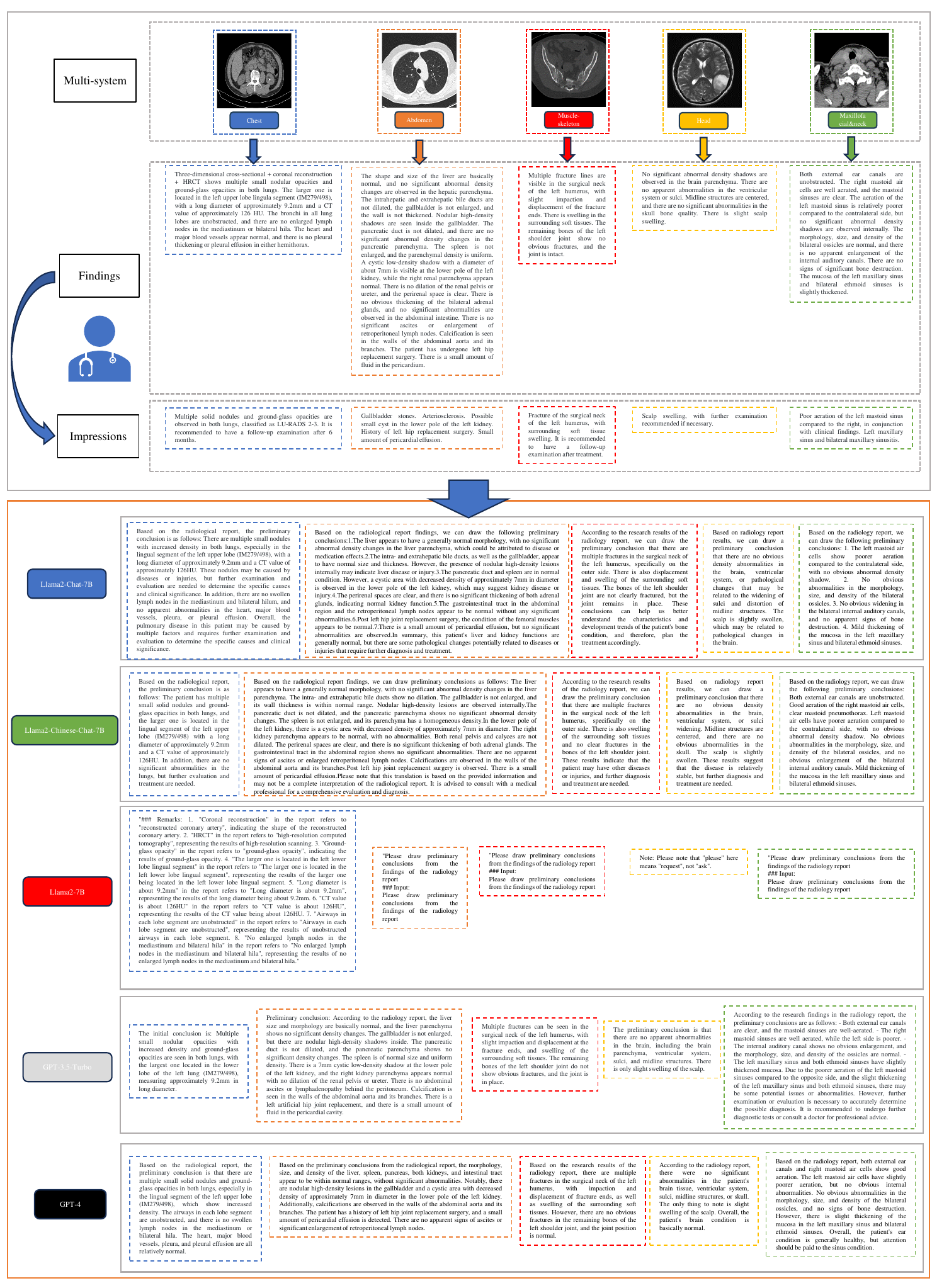}
    \caption{Diagram of the response results among fine-tuned models and state-of-the-art LLMs across systems (Part 2). Within each system (i.e., chest, abdomen, muscle-skeleton, head, and maxillofacial \& neck), each model generates corresponding clinical inference based on the same finding.}
    \label{fig:3.2-2}
\end{figure}

\begin{figure}[htbp!]
    \centering

    \includegraphics[width=0.87\textwidth,height=1.24\textwidth]{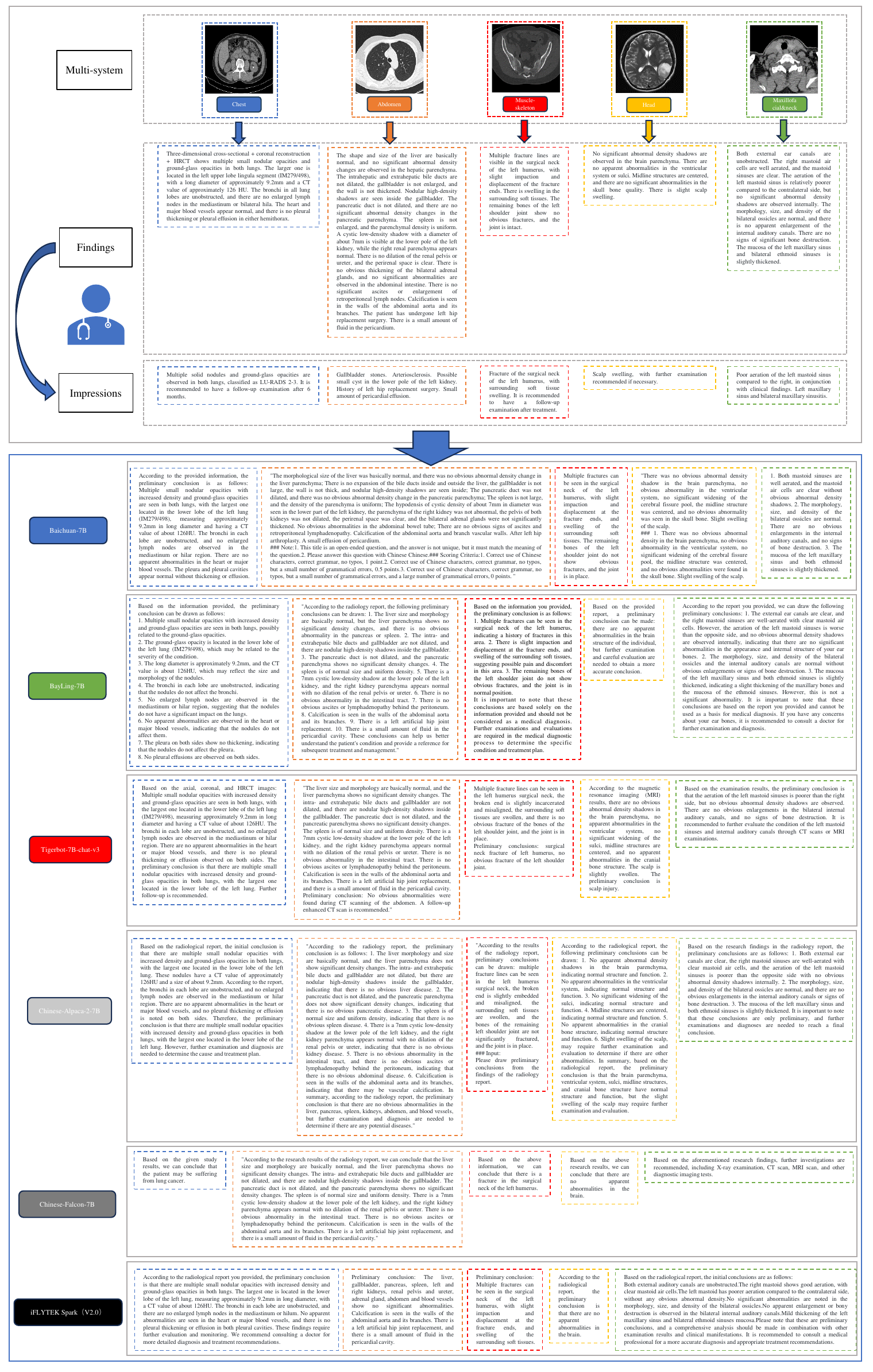}
    \caption{Diagram of the response results among fine-tuned models and state-of-the-art LLMs across systems (Part 3). Within each system (i.e., chest, abdomen, muscle-skeleton, head, and maxillofacial \& neck), each model generates corresponding clinical inference based on the same finding.}
    \label{fig:3.2-3}
\end{figure}

Supported by Figure \ref{fig:expertEval}, experts' evaluation on the seven clinical metrics, proposed in Section \ref{sec:Performance Evaluation Indexes} and Figure \ref{fig5}, have acknowledged the power of our fine-tuning strategy (i.e., all fine-tuned methods cover larger areas in each plot than others do, even beating the prevailing GPT family). Of greater significance, among all the fine-tuned models, our ChatRadio-Valuer consistently achieves the highest scores on these seven metrics in most cases (either in multi-institution evaluation or in multi-system evaluation), thereby demonstrating its substantial potential and value in clinical practice.

\section{Discussion}

\textbf{Significance:} This work demonstrates the potential of using large language models for automatic radiology report generation. By training a model on a large multi-institution, multi-system dataset, this method shows the feasibility of generating accurate and coherent impressions from radiology findings. The model's strong performance even when evaluated on external test sets indicates its ability to generalize across institutions and body systems. This has significant implications for improving radiologist workflow efficiency and standardizing impression styles across healthcare systems. More broadly, it highlights the promise of large language models for clinical NLP tasks given sufficient domain-specific training data.

\textbf{Insights:} While our proposed model outperforms the previous models and obtains a state-of-the-art performance, the subgroup analysis presents different performances in specific system radiology reports. In the current study, the performance of our proposed model in generating head and maxillofacial $\&$ neck CT reports is inferior to that of chest, abdominal, and musculoskeletal reports. This maybe due to the more comprehensive description of anatomical structure and multi-sequence of MRI reports, making it more difficult for the model to conclude and summarize the impression. Including more training data would be a strategy to improve the performance in these subgroups. Another reason may be that different radiologists have respective styles in drawing a conclusion for the same abnormities in head radiology reports. For example, the presenting spotted lesions with hypointensity in T1-weighted imaging, hyperintensity in T2-weighted imaging, and FLAIR imaging may be concluded as “ischemia lesions”, “cerebral white matter lesions” or “infraction lesions” by different radiologists. This also may confuse the model and limit its performance.

\textbf{Limitations:} Although this work systematically evaluates ChatRadio-Valuer for radiology report generation in detecting diverse diseases, there are several limitations and challenges requiring further exploration in future work. First, we only included Chinese radiology reports and the performance of our proposed model in other languages or styles of radiology reports should be further investigated with more diverse and balanced data distribution to verify the model generalization.
Second, considering only CT and MRI reports were collected and analyzed in the current study, other image modality reports (e.g., X-ray, PETCT, and ultrasound) will be our future direction. Third, reports generation based on causal explanatory reasoning is very necessary, which will guarantee and promote the reliability of the impressions from LLMs. These aspects will be paid more attention in our future work.

\textbf{Generalizability:} While tested on CT reports from Chinese hospitals, this approach could potentially generalize to other report types given sufficient training data. However, there are several factors affecting generalizability to other languages and healthcare systems. Radiology reporting conventions differ across countries, which could impact impression styles. Out-of-vocabulary medical terms may also emerge. Extensive testing and adaptation to localized datasets would be necessary before deployment in other global contexts. Furthermore, the model may need retraining if deployed in specialized settings like pediatric hospitals where disease patterns differ. Overall, the model shows promising generalization capacity within the Chinese clinical context, but global applicability would require additional investigation.

\section{Conclusion and Future Work}

In our work, we delve into the exploration current obstacles in clinical AI applications of radiology—notably, the complicated heterogeneity of multi-institution and multi-system data distribution, limited performance and generalizability. This avenue of research is of critical importance in the contemporary digital medical landscape, where radiology report generation plays a vital role yet are often complex and time-consuming to diagnose and annotate by the massive skilled experts.

To our knowledge, this investigation is groundbreaking in its focus as it offers fresh perspectives into the potential of employing large language models for the purpose of radiology report generation. Moreover, an innovative scheme is proposed in this paper, aiming for faster and more data-efficient diagnostics. This could revolutionize the diagnosis process for patients with systemic diseases.

We have verified the efficacy and efficiency of ChatRadio-Valuer in clinical-level domain adaptation scenarios, indicating high performance and generalizability for detecting diverse diseases based on radiology reports. Specifically, the experimental results reveal that the impressions generated by ChatRadio-Valuer are more prevailing than those of the state-of-the-art LLMs (such as ChatGPT and GPT-4 et al.), based on massive and complex radiology reports from multiple institutions and systems. The superiority is evident in terms of rigorous engineering indicators, clinical efficacy, and deployment cost metrics when evaluated by radiologists with varying levels of experience. It's also worth noting that many referring physicians favor the impressions generated by ChatRadio-Valuer, attributing this perceived superior preference to its better coherence and a lower tendency to omit vital information. This is particularly significant concerning metrics like missed diagnoses and overdiagnosis, which heavily impact on clinical utility.
 
Looking ahead to the future, the potential for integrating the text-handling capabilities of LLMs with other analysis methods appears substantial. Such synergistic combinations could unlock new approaches to radiology reports generation analysis, leading to broader medical generalizability across more languages styles and more modalities carriers.

\section*{Acknowledgments}
The study is supported by National Natural Science Foundation of China (31971288, U1801265); National Natural Science Foundation of China (82102157, U22A2034); Hunan Provincial Natural Science Foundation for Excellent Young Scholars (2022JJ20089); Research Project of Postgraduate Education and Teaching Reform of Central South University (2022JGB117); Clinical Research Center For Medical Imaging In Hunan Province (2020SK4001); Science and Technology Innovation Program of Hunan Province (2021RC4016); Central South University Research Programme of Advanced Interdisciplinary Studies (No. 2023QYJC020)



\bibliography{LLM_refs}
\bibliographystyle{unsrt}


\section*{\huge Appendix}

\end{document}